\documentclass[journal]{IEEEtran}
\usepackage{times}
\usepackage{lineno}
\usepackage{amsmath}
\usepackage{amssymb}
\usepackage{commath}
\usepackage{tabularx}
\usepackage{booktabs}
\usepackage{float}
\usepackage{algorithm}
\usepackage{algpseudocode}
\usepackage{graphicx}
\usepackage{subfigure}
\usepackage{color}
\usepackage{CJK}
\usepackage{cite}
\usepackage{multirow}

\begin{document}
\title{On Efficient and Robust Metrics for RANSAC Hypotheses and 3D Rigid Registration}
\author{Jiaqi~Yang, Zhiqiang~Huang, Siwen~Quan, Qian~Zhang,  Yanning~Zhang,~\IEEEmembership{Senior Member, IEEE} and Zhiguo~Cao
	
\thanks{This work is supported in part by the National Key R\&D Program of China (No. 2018YFB1305500), the Natural Science Basic Research Plan in Shaanxi Province of China (No. 2020JQ-210),  the National Natural Science Foundation of China (No. 41801274), and the Fundamental Research Funds for the Central Universities (No. D5000200078 \& 300102320304)
	
	\IEEEcompsocthanksitem Jiaqi~Yang and Yanning Zhang are with the National Engineering Laboratory for Integrated Aero-Space-Ground-Ocean Big Data Application Technology, School of Computer Science, Northwestern Polytechnical University, China. E-mail: jqyang@nwpu.edu.cn; ynzhang@nwpu.edu.cn.
	\IEEEcompsocthanksitem Zhiqiang Huang is with the School of Software, Northwestern Polytechnical University,  Xi'an 710129, China. E-mail: zhiqianghuang@mail.nwpu.edu.cn.
	\IEEEcompsocthanksitem Siwen Quan is with the School of Electronic and Control Engineering, Chang'an University, Xi'an 710064,	China. E-mail: siwenquan@chd.edu.cn.
	\IEEEcompsocthanksitem Qian Zhang is  with the School of Resources and Environment, Hubei University, Wuhan 430062, China. E-mail: hangfanzq@163.com.
	\IEEEcompsocthanksitem Zhiguo Cao is with the National Key Laboratory of Science and Technology on Multi-spectral Information Processing, School of Artificial Intelligence and Automation, Huazhong University of Science and Technology, Wuhan 430074, China. E-mail: zgcao@hust.edu.cn. (\textit{Corresponding author: Zhiguo Cao})
}
}
\markboth{Journal of \LaTeX\ Class Files,~Vol.~14, No.~8, August~2015}%
{Shell \MakeLowercase{\textit{et al.}}: Bare Demo of IEEEtran.cls for IEEE Journals}

\maketitle

\begin{abstract}
This paper focuses on developing efficient and robust evaluation metrics for RANSAC hypotheses to achieve accurate 3D rigid registration. Estimating six-degree-of-freedom (6-DoF) pose from feature correspondences remains a popular approach to 3D rigid registration, where random sample consensus (RANSAC) is a de-facto choice to this problem. However, existing metrics for RANSAC hypotheses are either time-consuming or sensitive to common nuisances, parameter variations, and different application scenarios, resulting in performance deterioration in overall registration accuracy and speed. We alleviate this problem by first analyzing the contributions of inliers and outliers, and then proposing several efficient and robust metrics with different designing motivations for RANSAC hypotheses. Comparative experiments on four standard datasets with different nuisances and application scenarios verify that the proposed metrics can significantly improve the registration performance and are more robust than several state-of-the-art competitors, making them good gifts to practical applications. This work also draws an interesting conclusion, i.e., not all inliers are equal while all outliers should be equal, which may shed new light on this research problem.
\end{abstract}


\IEEEpeerreviewmaketitle
\section{Introduction}\label{sec:intr}
\IEEEPARstart{R}{igid} registration of 3D point clouds is an active research area in computer vision with numerous applications such as 3D reconstruction~\cite{yang2018aligning,guo2014accurate}, 3D object recognition~\cite{guo20143d}, and localization~\cite{tateno20162}. The essential problem is estimating a  six-degree-of-freedom (6-DoF) pose to transform the source point cloud to the coordinate system of the target point cloud. Methods based on point-to-point feature correspondences, due to their robustness to occlusion, are widely employed in various registration scenarios~\cite{mian2005automatic,guo20143d,lei2017fast}. However, initial correspondences generated by matching local geometric descriptors are usually contaminated by heavy outliers owing to nuisances such as noise, varying data resolutions, clutter, and occlusion~\cite{Yang2020corr_group_eval}. Therefore, 6-DoF pose estimation from correspondences with low inlier ratios remains a challenging problem.

Random sample consensus (RANSAC)~\cite{fischler1981random} appears to be a de-facto solution to this problem. It iteratively generates 6-DoF pose hypotheses from correspondences and finds the best hypothesis as the final solution based on some hypothesis evaluation metrics. Although there have been a number of RANSAC variants~\cite{yang2016fast,yang2017multi,guo2015integrated}, they still follow the fundamental pipeline as RANSAC. Given the assumption that correct hypotheses are generated during the sampling process, the optimal solution can be found if with reasonable hypothesis evaluation metrics. Existing metrics for RANSAC hypotheses can be divided into two categories: correspondences-based and point-cloud-based. Metrics in the former category, e.g., inlier count~\cite{fischler1981random,rusu20113d} and Huber loss~\cite{rusu2009fast}, are known to be quite efficient because only sparse correspondence data are leveraged. However, they need very careful parameter tuning and are sensitive to many nuisances as will be verified Sect.~\ref{sec:exp}. Metrics in the latter category, e.g., the number of overlapped points~\cite{Quan2018Local} and point cloud distance~\cite{yang2016fast}, consider the whole point cloud for hypothesis evaluation. It has been demonstrated that these metrics can effectively improve the registration performance on some benchmarks. Nonetheless, they are particularly time-consuming because dense point clouds are treated as the input and nearest neighbor search is required. Therefore, existing hypothesis evaluation metrics for RANSAC in the context of 3D rigid registration fail to achieve a good balance between efficiency and robustness.
\\\\\noindent\textbf{Related work} This work focuses on the problem of defining efficient and robust hypothesis evaluation metrics for RANSAC to achieve accurate and efficient 3D rigid registration. We find that this problem has been more or less overlooked in the literature.

For RANSAC estimators applied to the 3D rigid registration problem,  the classical hypothesis evaluation metric, i.e., the inlier count, is arguably the most commonly employed one~\cite{mian2005automatic,rusu20113d}. In the recent decade, some other metrics have been proposed. Rusu et al.~\cite{rusu2009fast} proposed using the Huber loss as the evaluation metric for the sample consensus initial alignment (SAC-IA) estimator, i.e., a variant of RANSAC. Both inlier count and Huber loss only leverage sparse correspondence data. Yang et al.~\cite{yang2016fast} proposed the first point-cloud-based metric for RANSAC hypotheses named the point cloud distance, which is the mean of the distances of each point in the source point cloud to the target point cloud. Later, Quan et al.~\cite{Quan2018Local} also leveraged both point clouds during hypothesis evaluation but instead calculated the number of qualified points in the source point cloud that are close to the target point cloud.

Unfortunately, existing correspondences-based metrics are sensitive to a number of nuisances and parameter changes; point-cloud-based metrics are shown to be more robust but particularly time-consuming. In addition, we will show that different hypothesis evaluation metrics will result in significant registration performance variation. Although for RANSAC estimators applied to image registration, some metrics different from the classical inlier count metric have bee proposed~\cite{barath2018graph,lebeda2012fixing}, they fail to comprehensively analyze the contributions of inliers and outliers.
\\\\\noindent\textbf{Contributions} By reviewing existing related works, we find that the evaluation metric definition problem for RANSAC hypotheses has been overlooked and existing metrics fail to strike a good balance between robustness and efficiency. Consequently, we propose several continuous functions as hypothesis evaluation metrics, which evaluate the contributions of inliers and outliers in different ways. To show the impact of different hypothesis evaluation metrics on 3D registration performance and the effectiveness of the proposed metrics, comprehensive experiments on four datasets with comparison to all existing hypothesis evaluation metrics (to the best of our knowledge) have been conducted. The results suggest that our metrics can significantly improve the registration performance and are efficient as well as robust to common nuisances. In a nutshell, this paper has the following contributions.
\begin{itemize}
	\item  This paper proposes several metrics for RANSAC hypotheses, which evaluate the contributions of inliers and outliers with different motivations. The technique details are simple, while the performance boosting is impressive.
	\item In-depth experiments on four datasets with different application scenarios and nuisances have been conducted, to on one hand demonstrate the significance of hypothesis evaluation to the eventual 3D rigid registration performance and on the other show the overall superiority (i.e., fast, robust, and insensitive to parameter changes) of the proposed metrics. 
	\item We draw an interesting conclusion that {\textit{not all inliers are equal while all outliers should be equal}}, which may shed new light on this research problem.
\end{itemize}
\noindent\textbf{Paper organization} The remainder of this paper is structured as follows. Sect.~\ref{sec:mtd} introduces the technique details of the proposed metrics. Sect.~\ref{sec:exp} presents the experiments to validate the effectiveness of the proposed metrics with necessary explanations. Finally, Sect.~\ref{sec:conc} concludes the paper and presents potential future research directions.
\section{Methodology}\label{sec:mtd}
In this section, we first recap the RANSAC pipeline and introduce the role of hypothesis evaluation in this pipeline. Then, we analyze the contributions of inliers and outliers during hypothesis evaluation to guide metric definition. At last, several metrics for RANSAC hypotheses with different designing motivations are presented.
\subsection{Overview of RANSAC Pipeline}
\begin{figure}[t]
	\centering
	\includegraphics[width=1\linewidth]{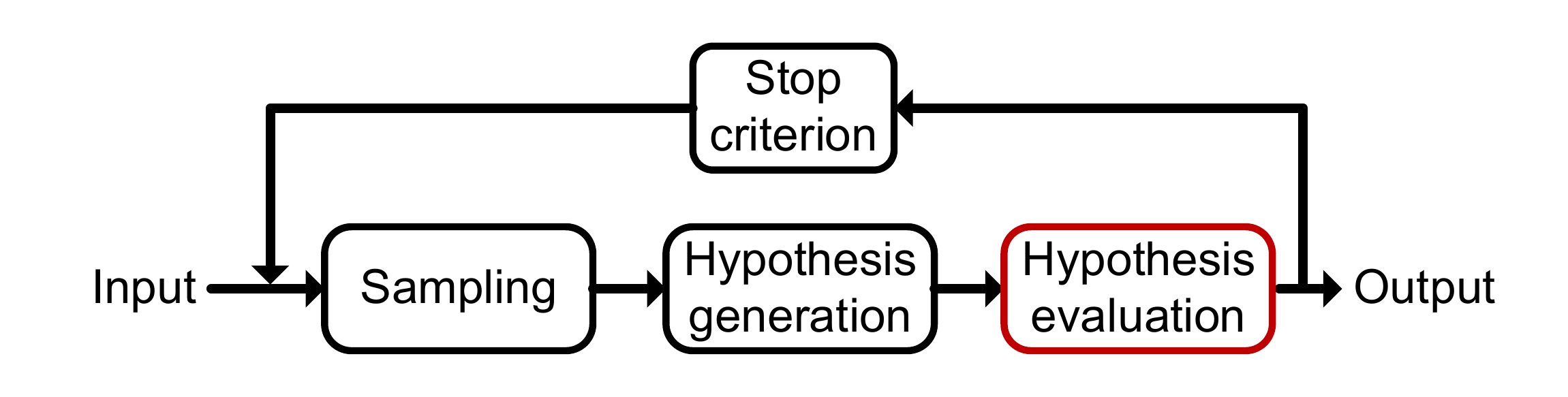}\\
	\caption{The general pipeline of RANSAC for 6-DoF pose estimation.}
	\label{fig:ransac_pipeline}
\end{figure}
As shown in Fig.~\ref{fig:ransac_pipeline}, RANSAC first iteratively samples correspondences from the initial correspondence set. Second, 6-DoF pose hypotheses are generated based on these samples with 1-point~\cite{guo2013rotational}, 2-point~\cite{yang2017multi}, or 3-point solutions~\cite{rusu2008aligning}. The problem then boils down to finding the best hypothesis from generated hypotheses, which is accomplished by the hypothesis evaluation stage. The iteration process stops if the number of repetitions reaches a pre-defined threshold or the best hypothesis from generated ones confirmed by a hypothesis evaluation metric is well enough. Finally, RANSAC outputs the best hypothesis.

Clearly, given the assumption that correct hypotheses are generated, it is critical for the overall algorithm to quickly find a correct one from generated hypotheses, which relies on an efficient and robust hypothesis evaluation metric. This is the focus of this work as well. Note that we will also test the effectiveness of the proposed metrics with different RANSAC sampling and hypothesis generation approaches.
\subsection{Analyzing the Contributions of Inliers and Outliers}\label{subsec:anal}
By taking correspondences-based and point-cloud-based hypothesis evaluation metrics into comparison, we find that metrics in the latter category are particularly time-consuming due to their {\textit{dense}} nature~\cite{yang2016fast}. Because hypothesis evaluation efficiency is one of the most critical issues in our designing motivation, we decide to devise correspondences-based metrics. We will first give a general definition to correspondences-based metrics, and then analyze the contributions of inliers and outliers.  

Let ${\bf P}^s$ and ${\bf P}^t$ be the source point cloud and the target point cloud, respectively. A correspondence in the given correspondence set $\bf C$ can be parametrized by ${\bf c}=({\bf p}^s,{\bf p}^t)$ with ${\bf p}^s \in {\bf P}^s$ and ${\bf p}^t \in {\bf P}^t$. Assume that at the $i$th iteration, a hypothesis ${\bf T}_i$ (comprised by a rotation pose ${\bf R}_i \in SO(3)$ and a translation pose ${\bf t}_i \in {\mathbb{R}^3}$) is generated. A  correspondences-based metric can be defined as:
\begin{equation}\label{eq:gen_form}
S({\bf T}_i) = \sum\limits_{j = 1}^n {s({{\bf c}_j})},
\end{equation}
where $n$ is the total number of correspondences in $\bf C$ and $s({{\bf c}_j})$ is a scoring function for a correspondence ${{\bf c}_j}$. In particular, $s({{\bf c}_j})$ can be represented as:
\begin{equation}\label{eq:para_t}
s({{\bf c}_j})=\left\{ {\begin{array}{*{20}{c}}
	{f^+({\bf c}_j),}&{{\rm{if }}~{e({\bf c}_j)} < t}\\
	{f^-({\bf c}_j),}&{{\rm{otherwise}}{\rm{}}}
	\end{array}} \right.
\end{equation}
where $e({\bf c}_j)=||{\bf R}_i{\bf p}_j^s+{\bf t}_i-{\bf p}_j^t||$ represents the transformation error of ${\bf c}_j$ and $t$ is a distance threshold to judge if ${\bf c}_j$ is an inlier. Existing correspondences-based metrics intend to {\bf 1)} define $f^+({\bf c}_j)$ as 1, i.e., all inliers are equal, and {\bf 2)} define $f^-({\bf c}_j)$ as a penalty function or 0, i.e., outliers give negative contributions or do not provide contributions.
\begin{figure}[t]
	\centering
	\includegraphics[width=1\linewidth]{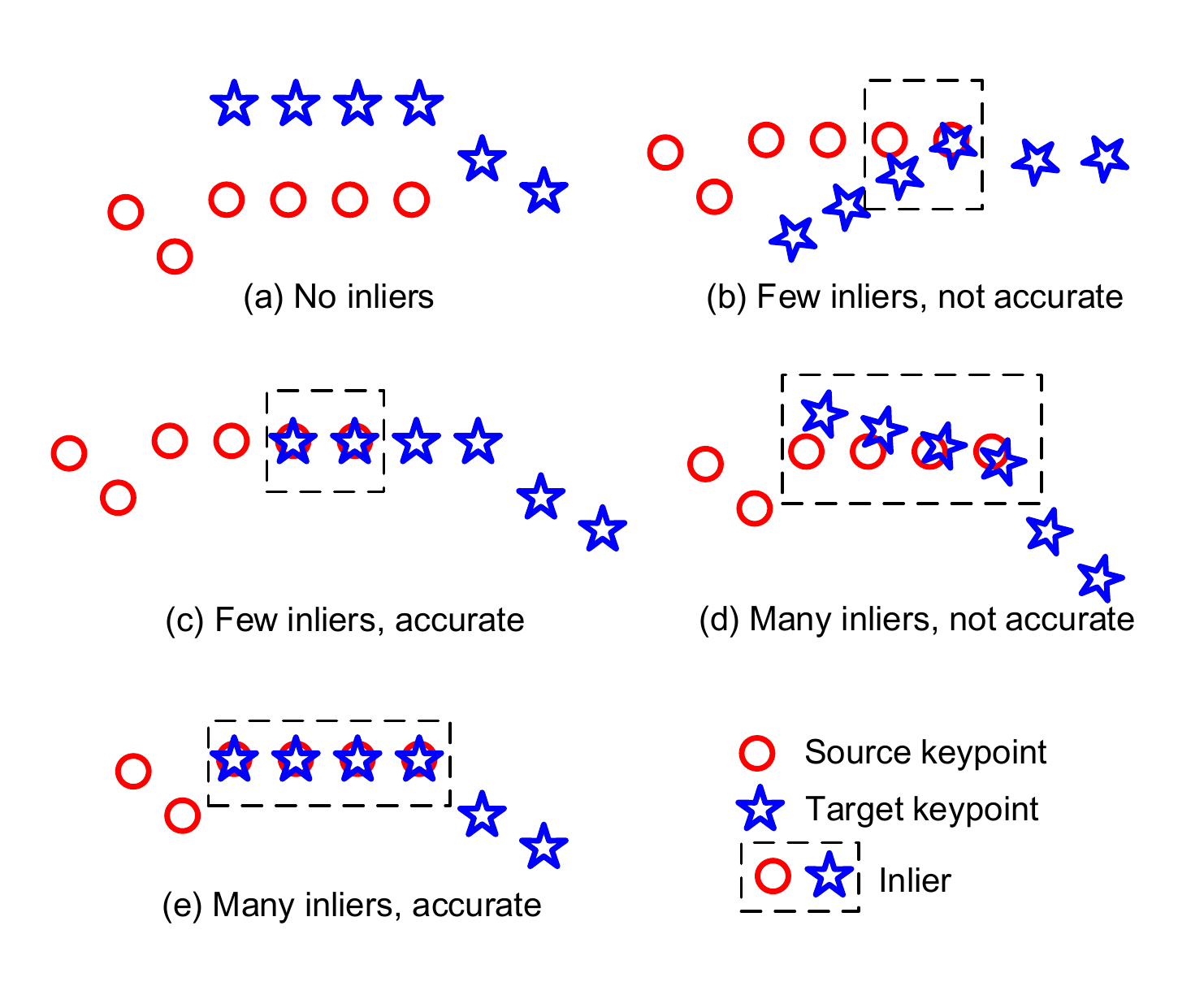}\\
	\caption{Five typical cases for hypothesis evaluation.}
	\label{fig:analyze_inlier_outlier}
\end{figure}

Unfortunately, we find that existing metrics fail to reasonably judge the contributions of inliers and outliers. As illustrated in Fig.~\ref{fig:analyze_inlier_outlier}, there are five typical cases for hypothesis evaluation. {\bf 1)} As for inliers, prior works~\cite{mian2005automatic,rusu2008aligning,yang2018aligning} mainly pursue as many as inliers and treat inliers equally. Nonetheless, although case (d) and (e) have the same inlier count, case (e) can result in more accurate registration. This suggests that {\textit{not all inliers are equal}. {\bf 2)} As for outliers, we suggest treating all outliers equally to {\textit{avoid their interferences to the contribution scores of inliers}}. This will be experimentally verified in Sect.~\ref{sec:exp}.
\begin{table*}[t]
	\centering
	\caption{Experimental datasets and their properties (SPC: source point cloud; TPS: target point cloud).}
	\label{tab:dataset}
	\scalebox{1}{
		\begin{tabular}{|c| c|c| c|c| c| c|}
			\hline
			\bf{Dataset} & \bf{Scenario}& \bf{Nuisances} & \bf{Modality}& \bf{\# SPC}& \bf{\# TPC}& \bf{\# Matching Pairs}\\
			\hline
			U3M~\cite{mian2006novel} & Registration & Limited overlap, self-occlusion & LiDAR & -- & -- & 496\\
			\hline
			BMR~\cite{salti2014shot} & Registration & Limited overlap, self-occlusion, real noise, holes & Kinect & -- & --  & 485\\
			\hline
			U3OR~\cite{mian2006three,mian2010repeatability} & Object recognition & Clutter, occlusion & LiDAR & 5 & 50 & 188\\
			\hline
			BoD5~\cite{salti2014shot}& Object recognition & Clutter, occlusion, real noise, holes & Kinect & 26 & 15 & 43\\
			\hline
		\end{tabular}}
	\end{table*}
\subsection{Proposed Metrics for RANSAC Hypotheses}
To ensure that hypothesis metrics serve case (e) in Fig.~\ref{fig:analyze_inlier_outlier} as the optimal case, we propose the following metrics with their plots being shown in Fig.~\ref{fig:metric_curve}.
\\\\\noindent\textbf{MAE} The mean absolute error (MAE) metric is used to measure the contribution of inliers.
\begin{equation}
s({{\bf c}_j})=\left\{ {\begin{array}{*{20}{c}}
	{\frac{{|e({{\bf{c}}_j}) - t|}}{t},}&{{\rm{if }}~{e({\bf c}_j)} < t}\\
	{0,}&{{\rm{otherwise}}{\rm{}}}
	\end{array}} \right.
\end{equation}
\\\\\noindent\textbf{MSE} The mean squared error (MSE) metric is employed to measure the contribution of inliers.
\begin{equation}
s({{\bf c}_j})=\left\{ {\begin{array}{*{20}{c}}
	{\frac{{|e({{\bf{c}}_j}) - t|^2}}{t^2},}&{{\rm{if }}~{e({\bf c}_j)} < t}\\
	{0,}&{{\rm{otherwise}}{\rm{}}}
	\end{array}} \right.
\end{equation}
\\\\\noindent\textbf{LOG-COSH} The LOG-COSH metric is used to measure the contribution of inliers.
\begin{equation}
s({{\bf c}_j})=\left\{ {\begin{array}{*{20}{c}}
	{\frac{{\log \left[ {\cosh (e({{\bf{c}}_j}) - t)} \right]}}{{\log \left[ {\cosh (t)} \right]}},}&{{\rm{if }}~{e({\bf c}_j)} < t}\\
	{0,}&{{\rm{otherwise}}{\rm{}}}
	\end{array}} \right.
\end{equation}
\\\\\noindent\textbf{EXP} The exponential (EXP) metric is considered to measure the contribution of inliers.
\begin{equation}
s({{\bf c}_j})=\left\{ {\begin{array}{*{20}{c}}
	{\exp \left( { - \frac{{e{{({{\bf{c}}_j})}^2}}}{{2{t^2}}}} \right),}&{{\rm{if }}~{e({\bf c}_j)} < t}\\
	{0,}&{{\rm{otherwise}}{\rm{}}}
	\end{array}} \right.
\end{equation}
\\\\\noindent\textbf{QUANTILE} The QUANTILE metric is proposed to measure the contribution of inliers and outliers.
\begin{equation}\label{eq:quant}
s({{\bf c}_j})=\left\{ {\begin{array}{*{20}{c}}
	{\frac{{m|e({{\bf{c}}_j}) - t|}}{t},}&{{\rm{if }}~{e({\bf c}_j)} < t}\\
	{\frac{{(1 - m)|e({{\bf{c}}_j}) - t|}}{{e({{\bf{c}}_j})}},}&{{\rm{otherwise}}{\rm{}}}
	\end{array}} \right.
\end{equation}
\\\\\noindent\textbf{-QUANTILE} A modification of the QUANTILE metric (we dub it as -QUANTILE) is proposed to measure the contribution of inliers and outliers.
\begin{equation}\label{eq:-quant}
s({{\bf c}_j})=\left\{ {\begin{array}{*{20}{c}}
	{\frac{{m|e({{\bf{c}}_j}) - t|}}{t},}&{{\rm{if }}~{e({\bf c}_j)} < t}\\
	{\frac{{(m - 1)|e({{\bf{c}}_j}) - t|}}{{e({{\bf{c}}_j})}},}&{{\rm{otherwise}}{\rm{}}}
	\end{array}} \right.
\end{equation}
\begin{figure}[t]
	\centering
	\includegraphics[width=1\linewidth]{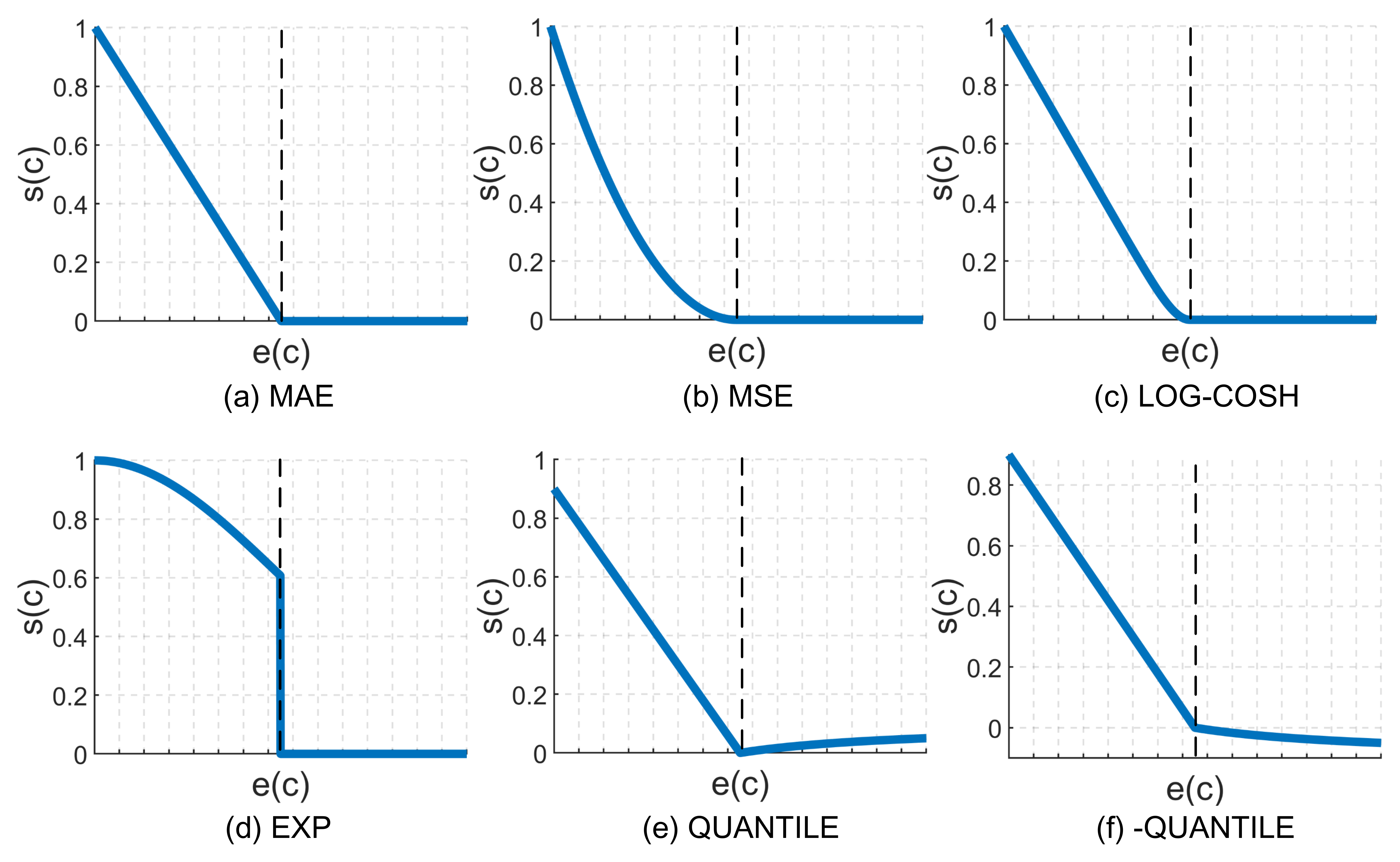}\\
	\caption{The general pipeline of RANSAC for 6-DoF pose estimation.}
	\label{fig:metric_curve}
\end{figure}
Several points should be noted here.
\begin{itemize}
	\item  The MAE, MSE, LOG-COSH, EXP metrics with $f^+({\bf c}_j)$ being continuous functions are proposed to opt more accurate inliers (with smaller $e({{\bf{c}}_j})$ values) and treat all outliers equally ($f^-({\bf c}_j)$ is set to 0). In machine learning areas, these metrics have been frequently employed as loss functions to tackle different problems. Specifically, we consider all of them in the 3D rigid registration problem to comprehensively evaluate the effects of different continuous functions on hypothesis evaluation and 3D rigid registration.
	\item The QUANTILE and -QUANTILE metrics are designed to support our statement that {\textit{all outliers should be equal}}. These two metrics assign positive and negative weights to outliers, respectively.
	\item In Eqs.~\ref{eq:quant} and~\ref{eq:-quant}, $m$ is a free parameter. We empirically set $m$ to 0.9 because we find that the conclusion to the contributions of outliers do no alter as long as $m \in (0,1)$.
\end{itemize}

We will use above metrics to assess the quality of a 6-DoF pose hypothesis ${\bf T}_i$. The hypothesis ${\bf T}^\star$ yielding the maximum scoring value (Eq.~\ref{eq:gen_form}) is served as the result of RANSAC. Two point clouds ${\bf P}^s$ and ${\bf P}^t$ can be aligned based on ${\bf P}^s\to{\bf P}'^s: {\bf p}'^s={\bf R}^\star{\bf p}^s+{\bf t}^\star, {\bf p}^s\in {\bf P}^s$, where ${\bf P}'^s$ is the transformed source point cloud after registration.
\section{Experiments}\label{sec:exp}
This section presents a comprehensive evaluation for the proposed metrics. In addition, we have compared our metrics with all existing hypothesis evaluation metrics (to the best of our knowledge) for RANSAC in 3D rigid registration scenarios. 

\subsection{Experimental Setup}
\subsubsection{Datasets}
\begin{figure}[t]
	\centering
	\includegraphics[width=1\linewidth]{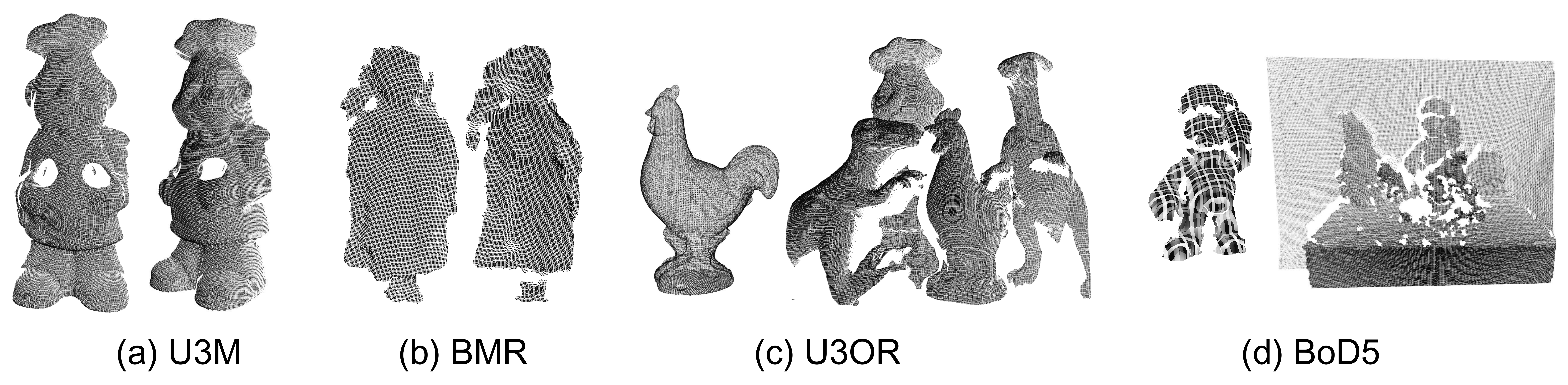}\\
	\caption{Sample views of point cloud pairs from experimental datasets.}
	\label{fig:dataset}
\end{figure}

Four standard datasets including  UWA 3D modeling (U3M)~\cite{mian2006novel}, Bologna Mesh Registration (BMR)~\cite{salti2014shot}, UWA 3D object recognition (U3OR)~\cite{mian2006three,mian2010repeatability}, and Bologna Dataset5 (BoD5)~\cite{salti2014shot} are considered in our experiments, as shown in Fig.~\ref{fig:dataset}. The main properties of experimental datasets are summarized in Table~\ref{tab:dataset}. One can see that they have {\bf 1)} different application scenarios, {\bf 2)} a variety of nuisances, and {\bf 3)} different data modalities, which can generate initial correspondence sets with a variety of {\textit {inlier ratios}}, {\textit {spatial distributions}}, and {\textit {scales}} to ensure a comprehensive evaluation.
\begin{figure}[t]
	\begin{minipage}{0.495\linewidth}
		\centering
		\subfigure[U3M]{
			\includegraphics[width=1\linewidth]{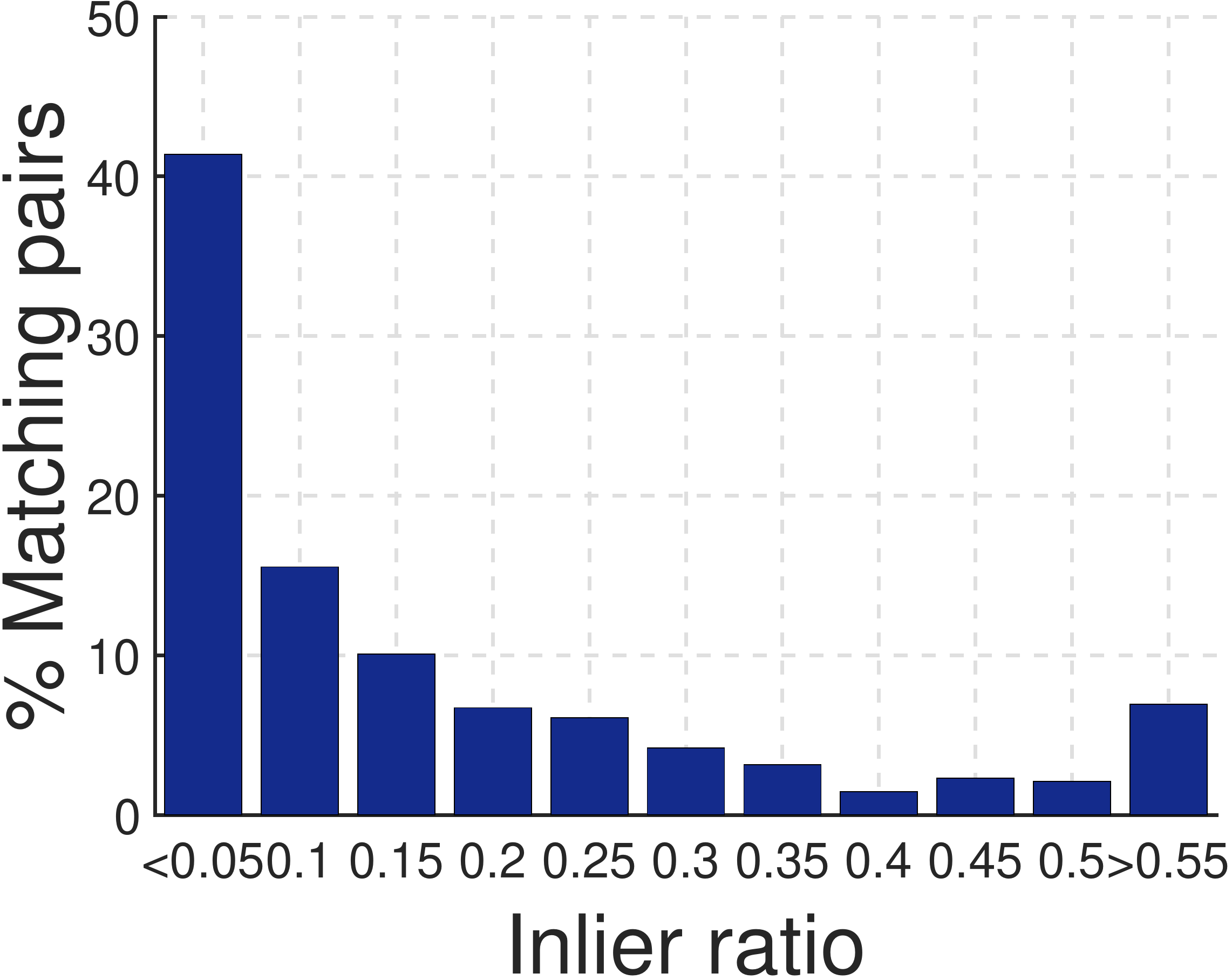}}
	\end{minipage}
	\begin{minipage}{0.495\linewidth}
		\centering
		\subfigure[BMR]{
			\includegraphics[width=1\linewidth]{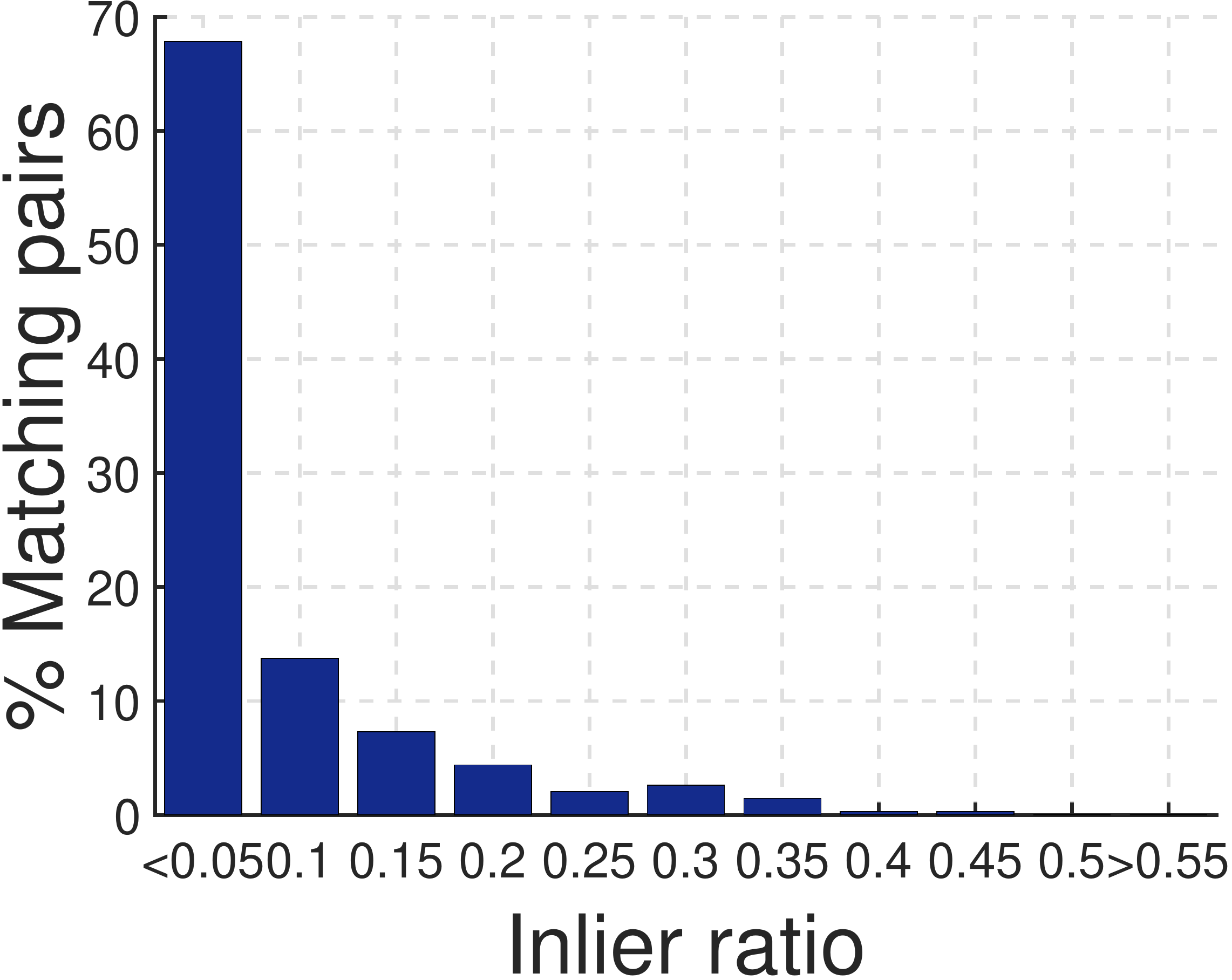}}
	\end{minipage}
	\begin{minipage}{0.495\linewidth}
		\centering
		\subfigure[U3OR]{
			\includegraphics[width=1\linewidth]{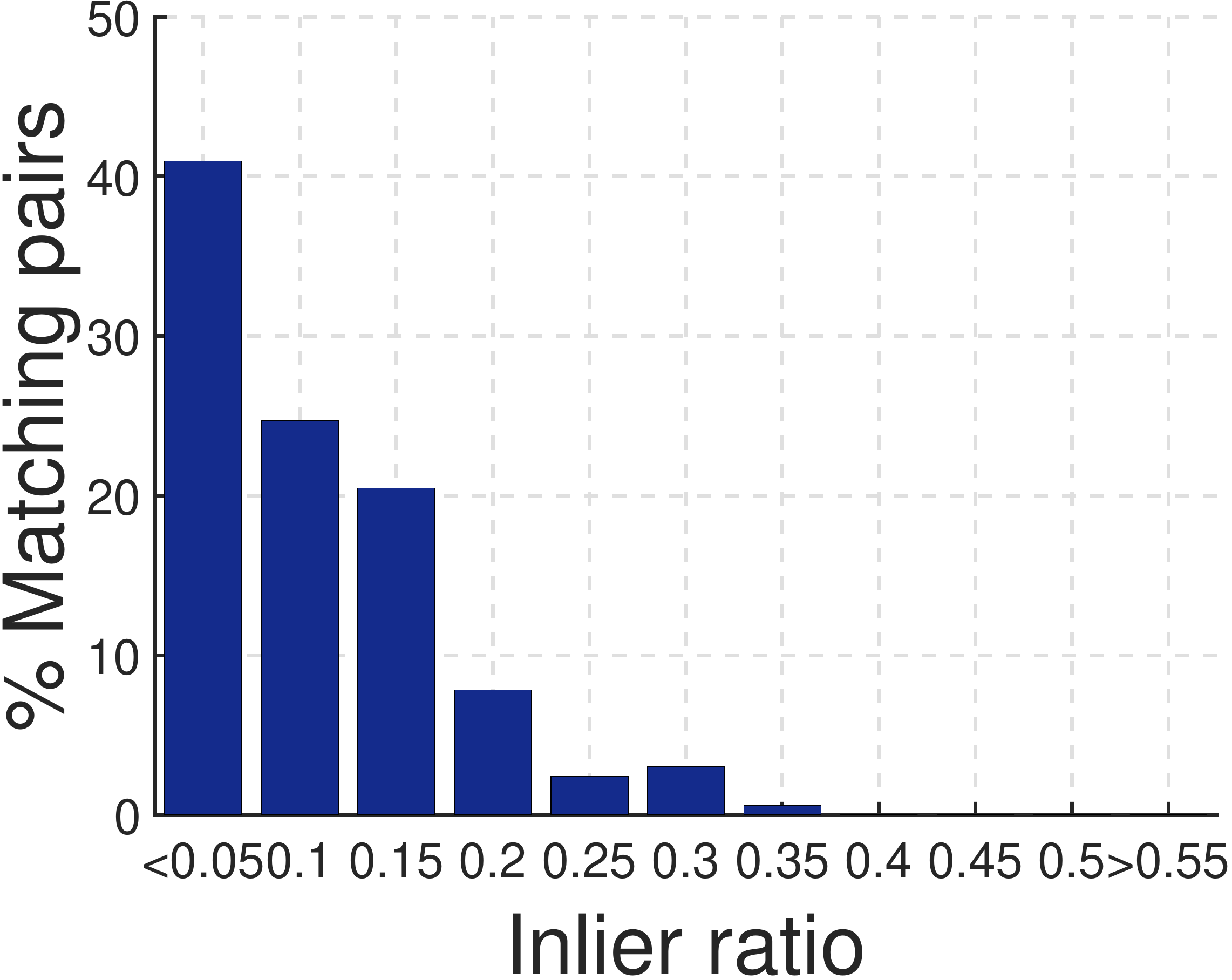}}
	\end{minipage}
	\begin{minipage}{0.495\linewidth}
		\centering
		\subfigure[BoD5]{
			\includegraphics[width=1\linewidth]{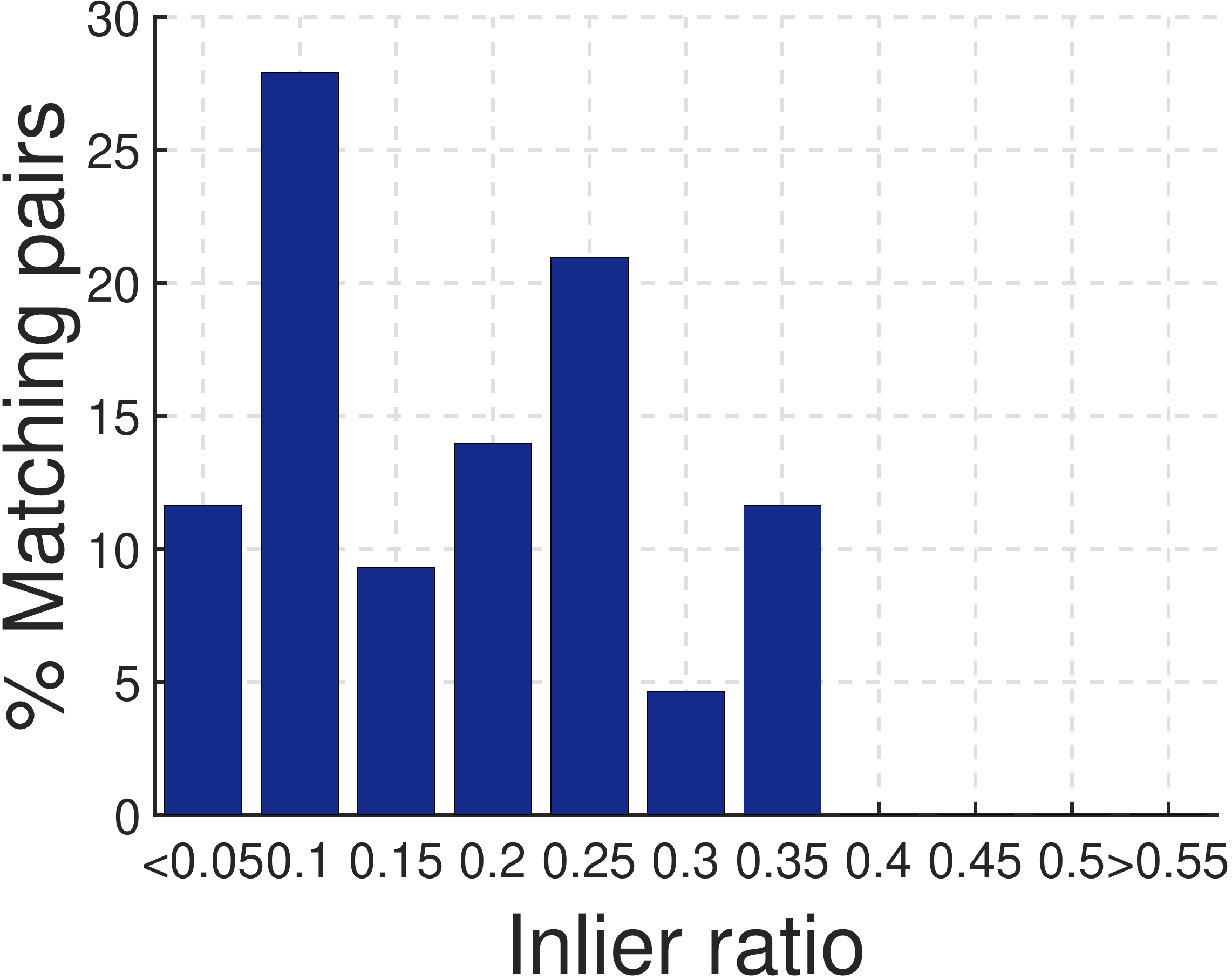}}
	\end{minipage}
	\caption{Inlier information of four experimental datasets.}
	\label{fig:dataset_info}
\end{figure}
\begin{figure}[htbp]
	\centering
	\includegraphics[width=1\linewidth]{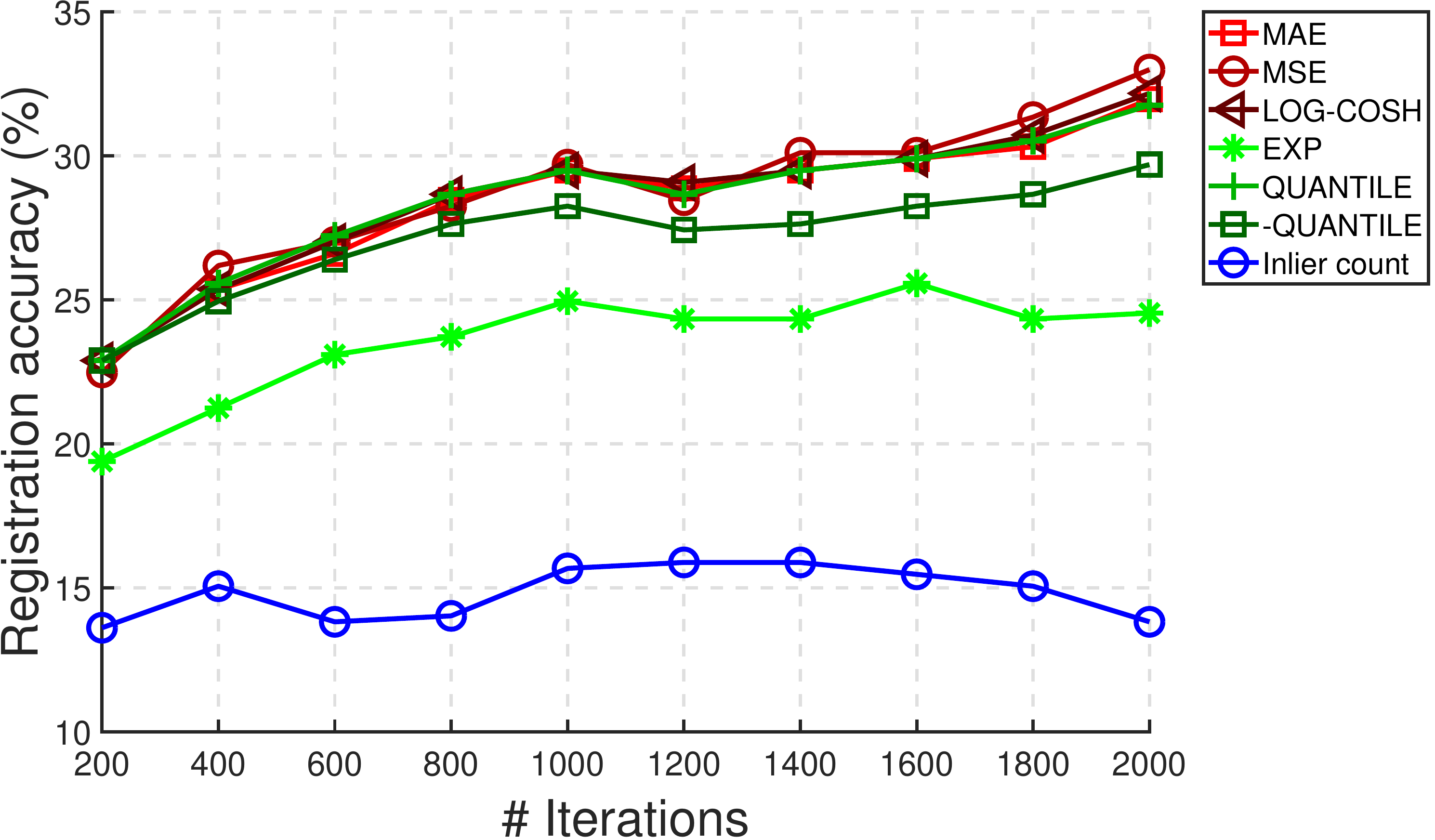}\\
	\caption{The effects of varying the number of iterations on registration accuracy.}
	\label{fig:para_iter}
\end{figure}
\subsubsection{Evaluation Criterion}
We use the root mean square error (RMSE)~\cite{choi2015robust,yang2016fast,zhou2016fast} criterion to assess the quality of a 3D rigid registration. Assume that ${\bf R}_{gt}$ and ${\bf t}_{gt}$ are the ground-truth rotation matrix and translation vector, respectively. The point-wise error $\epsilon$ between two truly corresponding points ${\bf p}^s$ and ${\bf p}^t$ is defined as:
\begin{equation}\label{eq:pnt_wise_err}
\begin{aligned}
\epsilon({\bf p}^s,{\bf p}^t)=\|{\bf R}_{gt}{\bf p}^s+ {\bf t}_{gt}-{\bf p}^t\|.
\end{aligned}
\end{equation}
RMSE is defined as:
\begin{equation}
\begin{aligned}
\rm{RMSE}=\sum\limits_{({\bf p}^s,{\bf p}^t)\in {\bf C}_{gt}} {\frac{\epsilon({\bf p}^s,{\bf p}^t)}{|{\bf C}_{gt}|}},
\end{aligned}
\end{equation}
where $ {\bf C}_{gt}$ is the ground truth set of corresponding points between the source point cloud and the target point cloud. A registration is judged as correct if its RMSE value is smaller than a distance threshold $d_{rmse}$. We define the registration accuracy of a method on a dataset as the ratio of correct registrations to the total number of point clouds to be registered in the dataset. 
\subsubsection{Implementation Details}
Inputs to RANSAC estimators for 3D rigid registration are feature correspondences. By default, we employ the Harris 3D (H3D)~\cite{sipiran2011harris} keypoint detector and the signatures of histograms of orientations (SHOT)~\cite{tombari2010unique} descriptor to detect keypoints on point clouds and describe local geometric features around keypoints, respectively. Then, feature correspondences between two point clouds are generated by matching local keypoint descriptors. Following Lowe's ratio rule~\cite{lowe2004distinctive}, 30\%$\times {\bf C}$ ($\bf C$ being the initial correspondence set) top-ranked feature correspondences in terms of Lowe's ratio score are selected as inputs to RANSAC estimators. Fig.~\ref{fig:dataset_info} visualizes the inlier ratio information of each dataset. We will also try different detector-descriptor combinations in Sect.~\ref{subsubsec:diff_conf}.

\begin{figure*}[t]
	\begin{minipage}{0.246\linewidth}
		\centering
		\subfigure[U3M]{
			\includegraphics[width=1\linewidth]{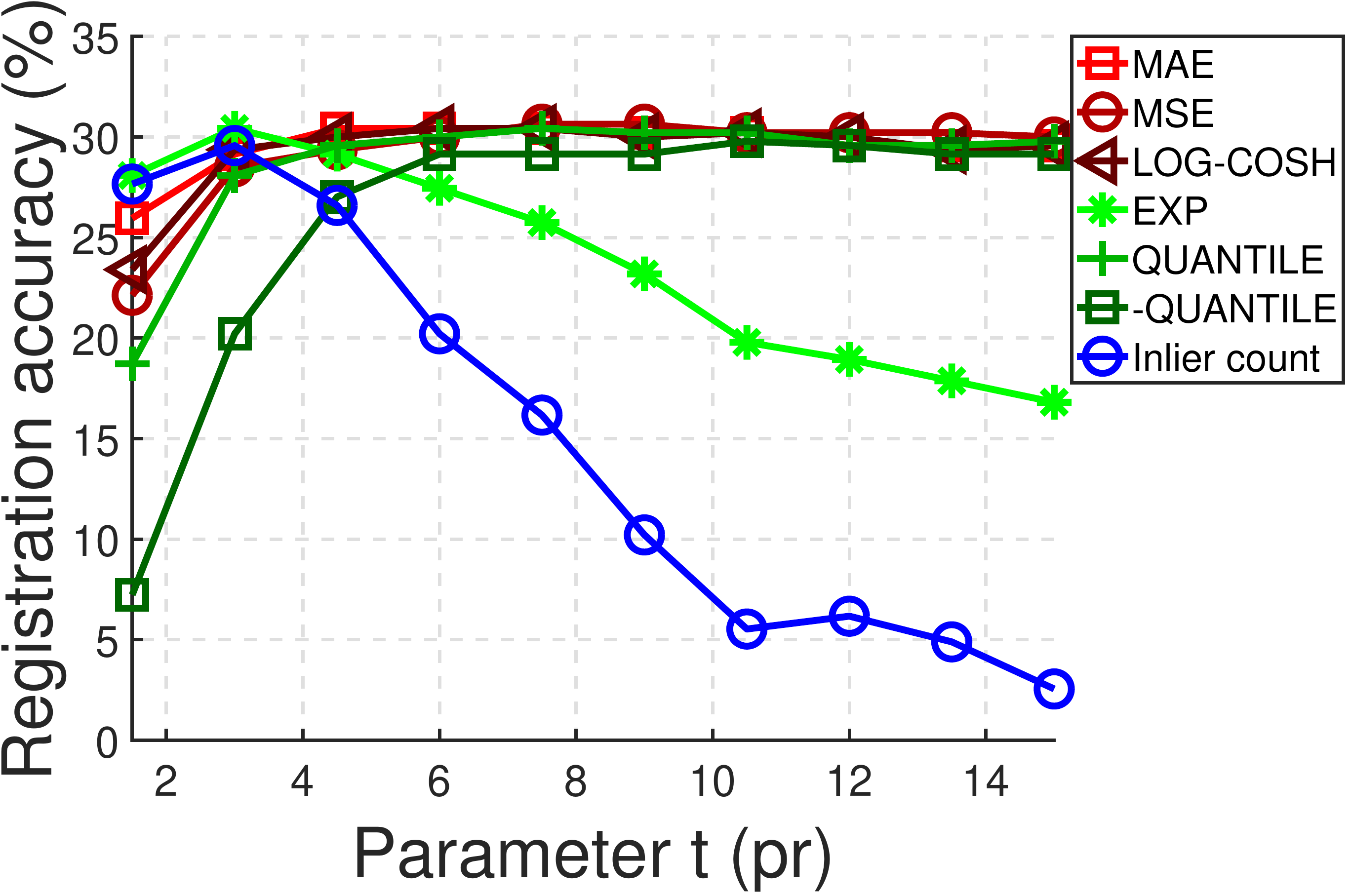}}
	\end{minipage}
	\begin{minipage}{0.246\linewidth}
		\centering
		\subfigure[BMR]{
			\includegraphics[width=1\linewidth]{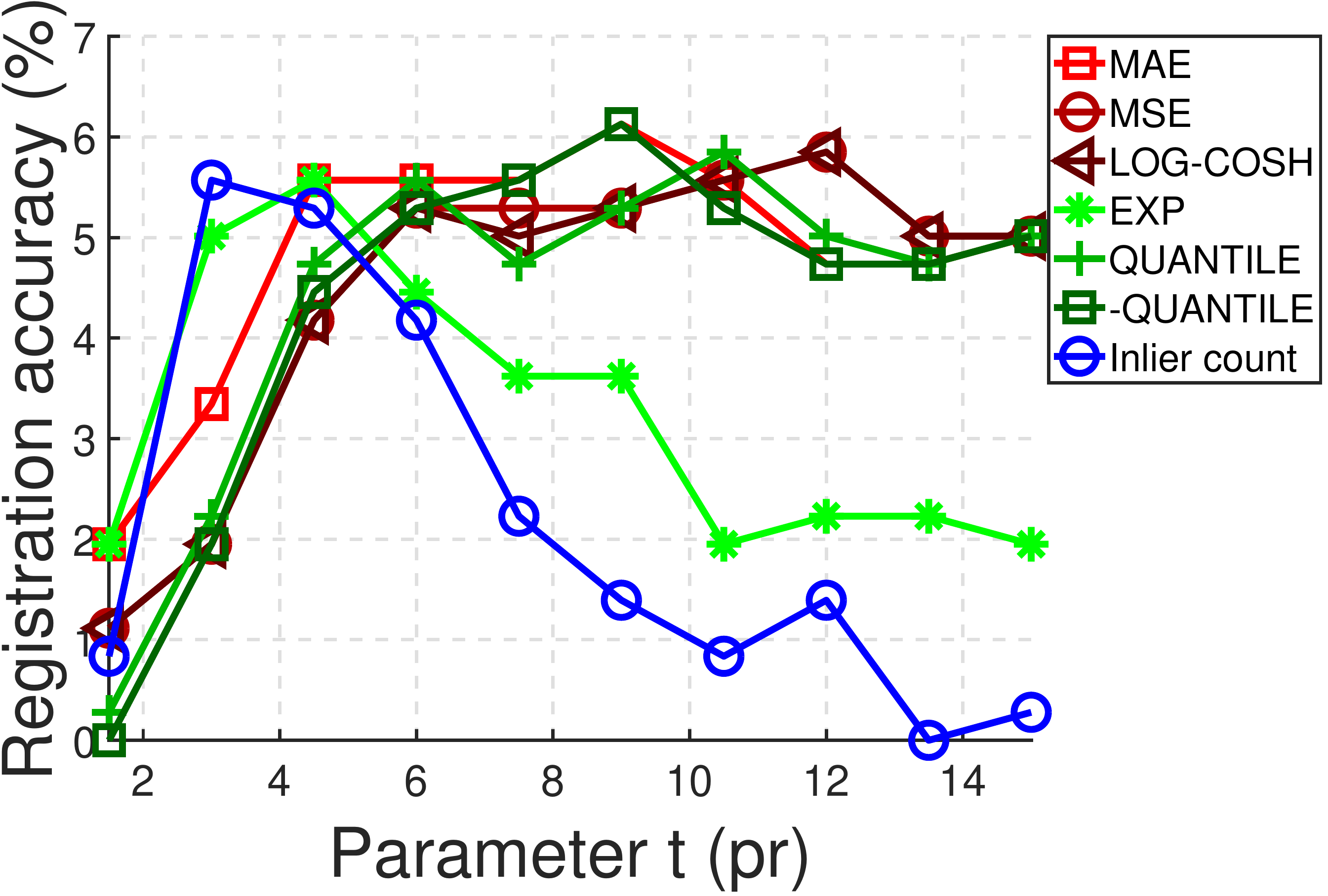}}
	\end{minipage}
	\begin{minipage}{0.246\linewidth}
		\centering
		\subfigure[U3OR]{
			\includegraphics[width=1\linewidth]{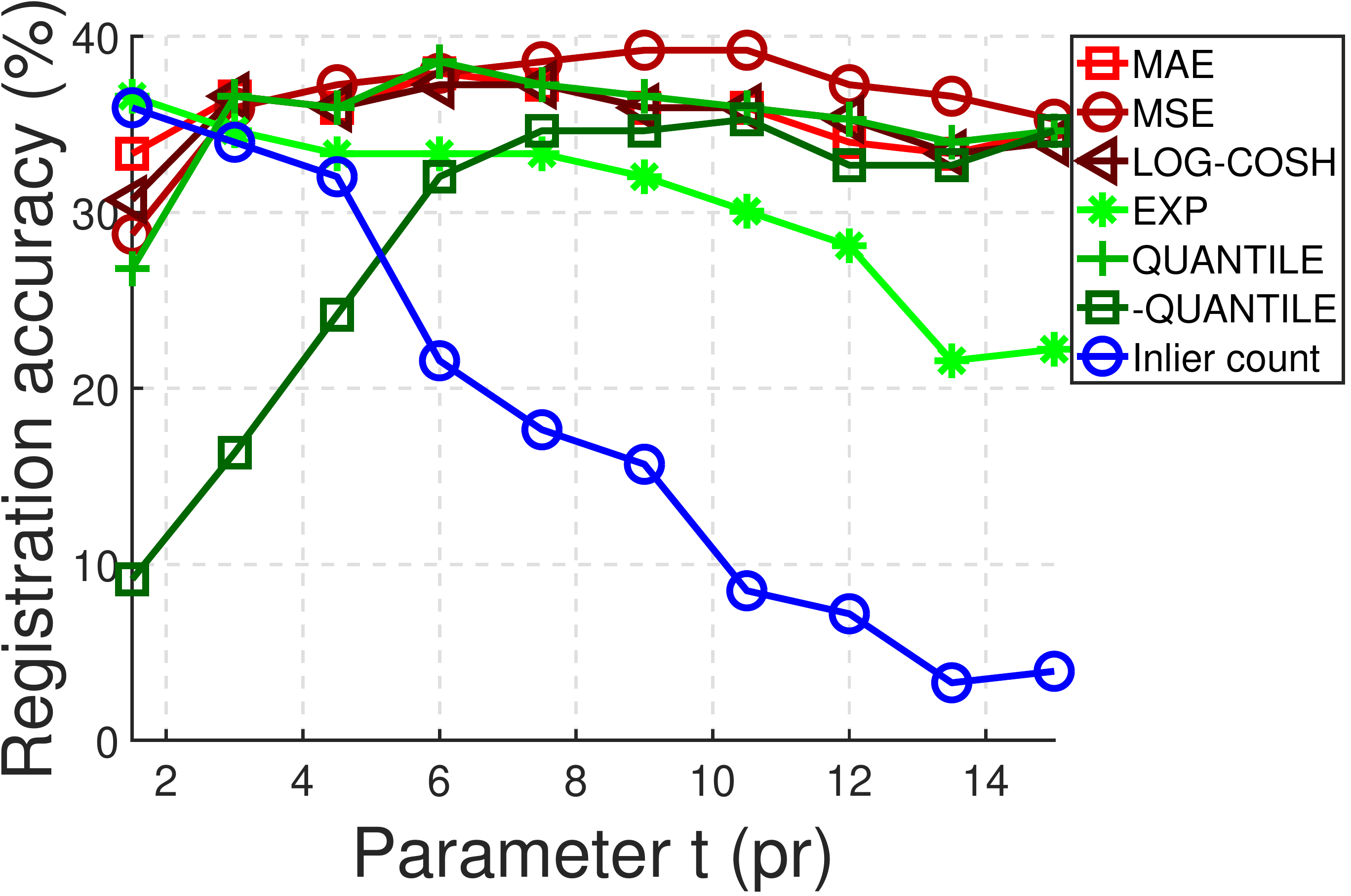}}
	\end{minipage}
	\begin{minipage}{0.246\linewidth}
		\centering
		\subfigure[BoD5]{
			\includegraphics[width=1\linewidth]{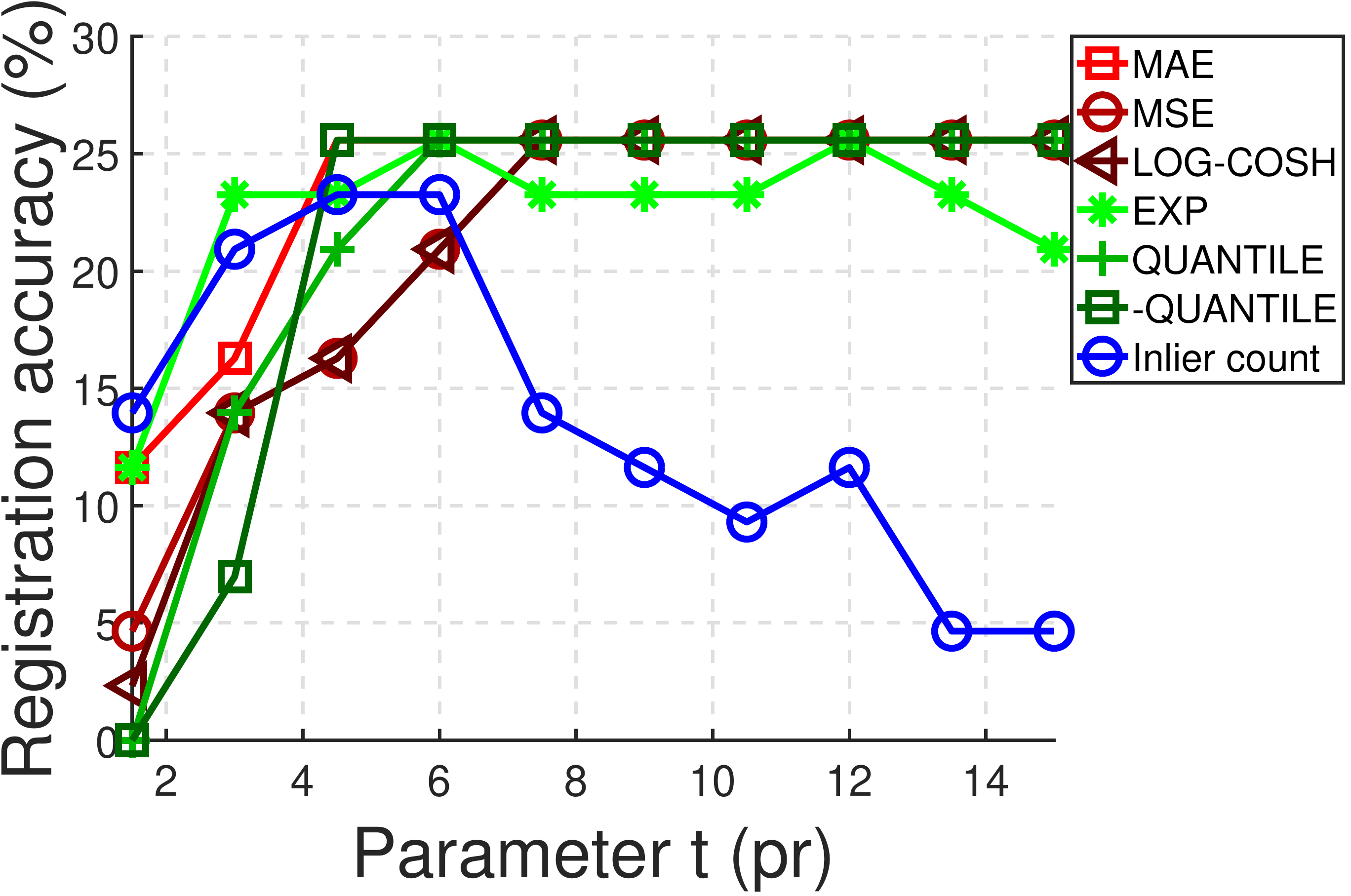}}
	\end{minipage}
	\caption{Sensitivity of tested metrics to parameter $t$ when varying datasets.}
	\label{fig:para_r_dataset}
\end{figure*}
\begin{figure*}[t]
	\begin{minipage}{0.246\linewidth}
		\centering
		\subfigure[$d_{rmse}$=0.5 pr]{
			\includegraphics[width=1\linewidth]{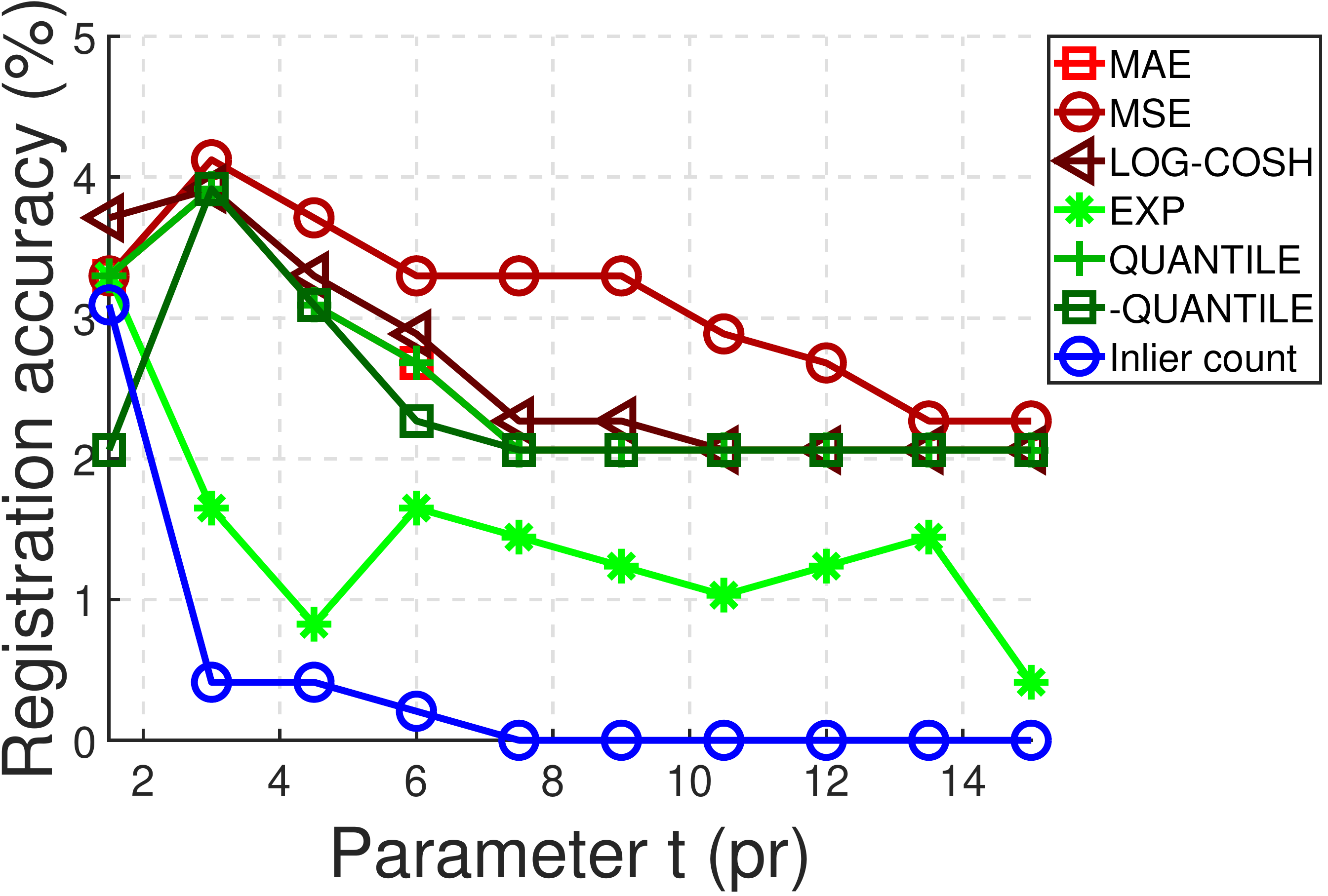}}
	\end{minipage}
	\begin{minipage}{0.246\linewidth}
		\centering
		\subfigure[$d_{rmse}$=1.0 pr]{
			\includegraphics[width=1\linewidth]{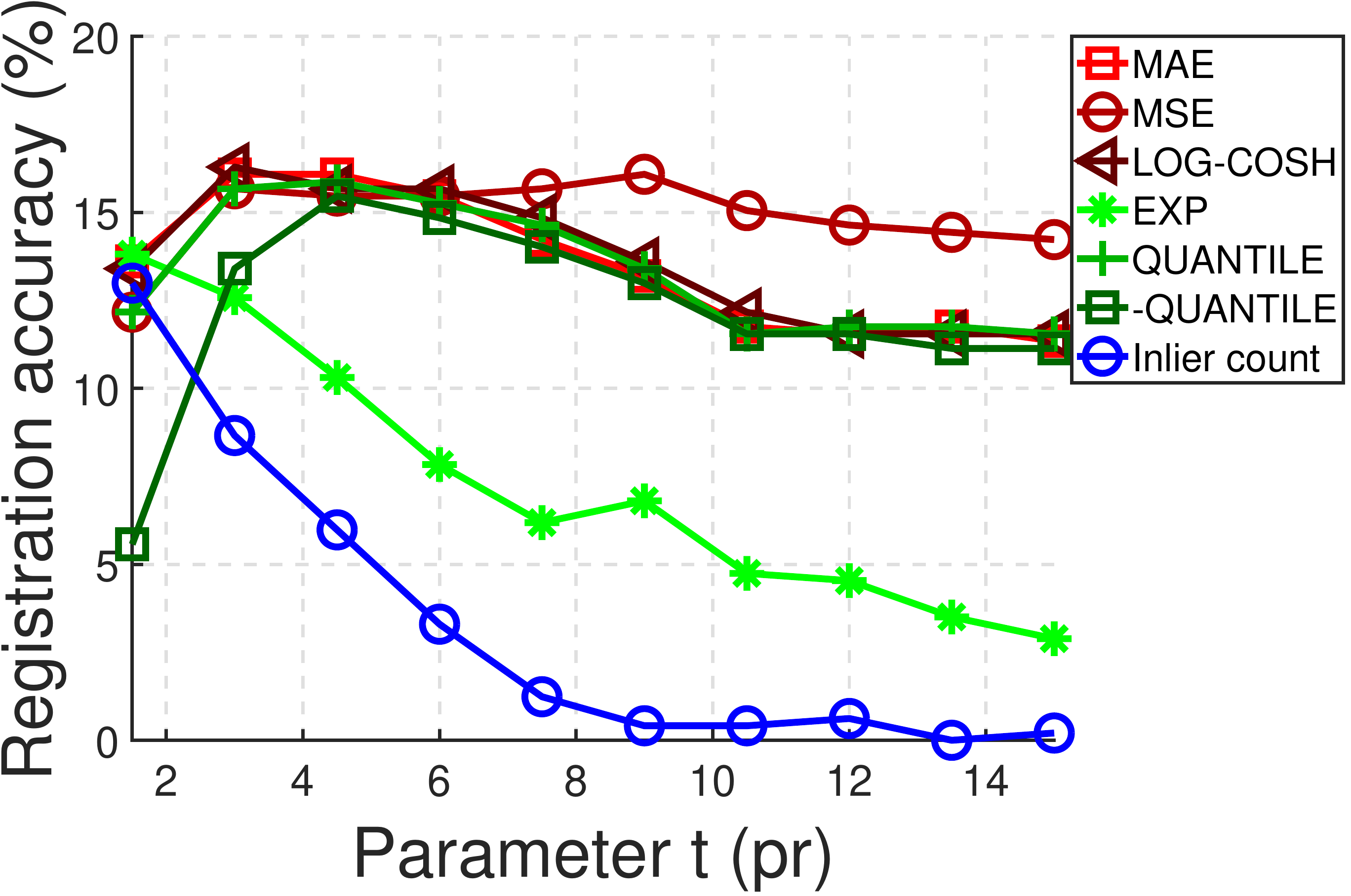}}
	\end{minipage}
	\begin{minipage}{0.246\linewidth}
		\centering
		\subfigure[$d_{rmse}$=1.5 pr]{
			\includegraphics[width=1\linewidth]{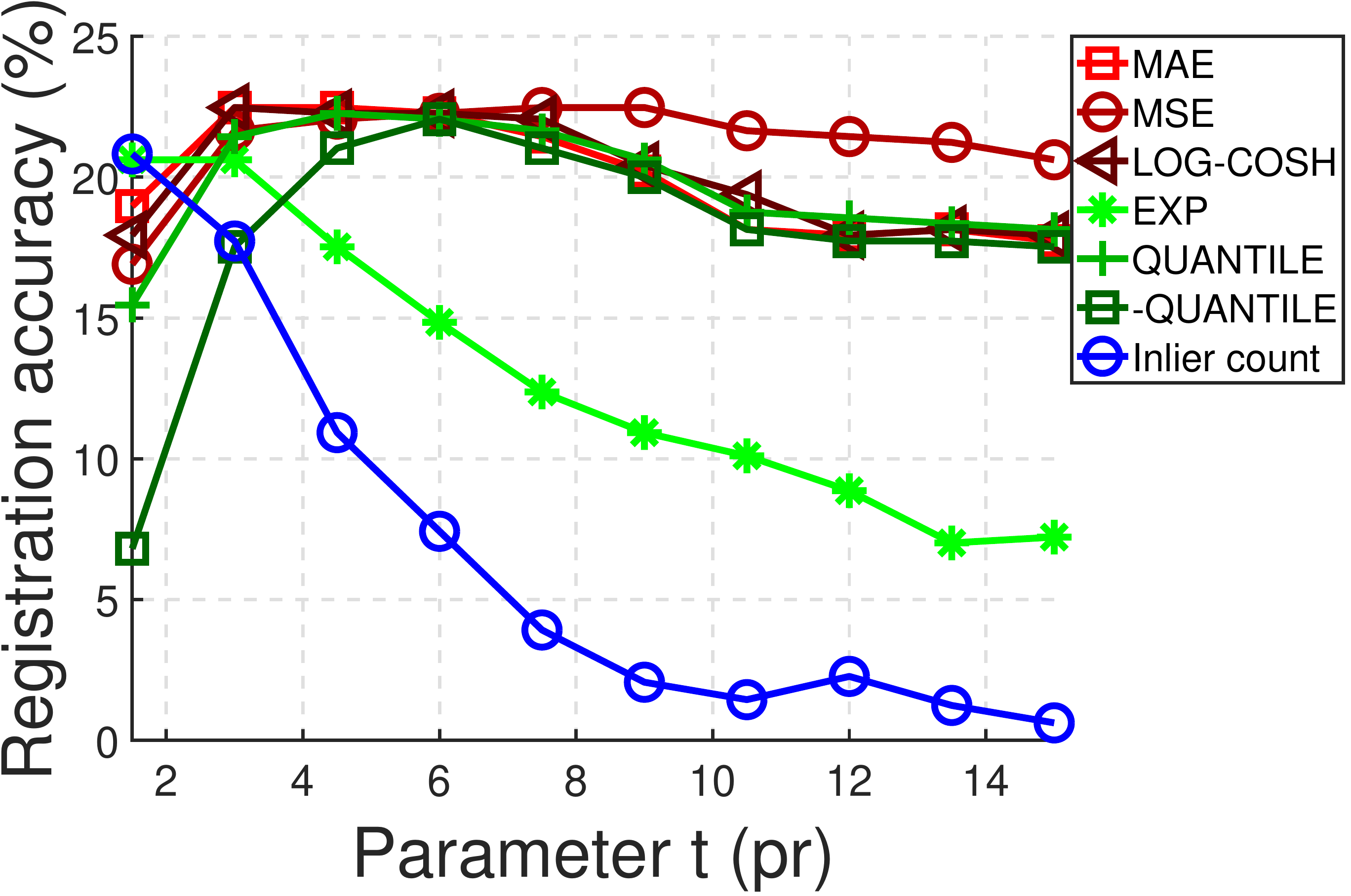}}
	\end{minipage}
	\begin{minipage}{0.246\linewidth}
		\centering
		\subfigure[$d_{rmse}$=2.0 pr]{
			\includegraphics[width=1\linewidth]{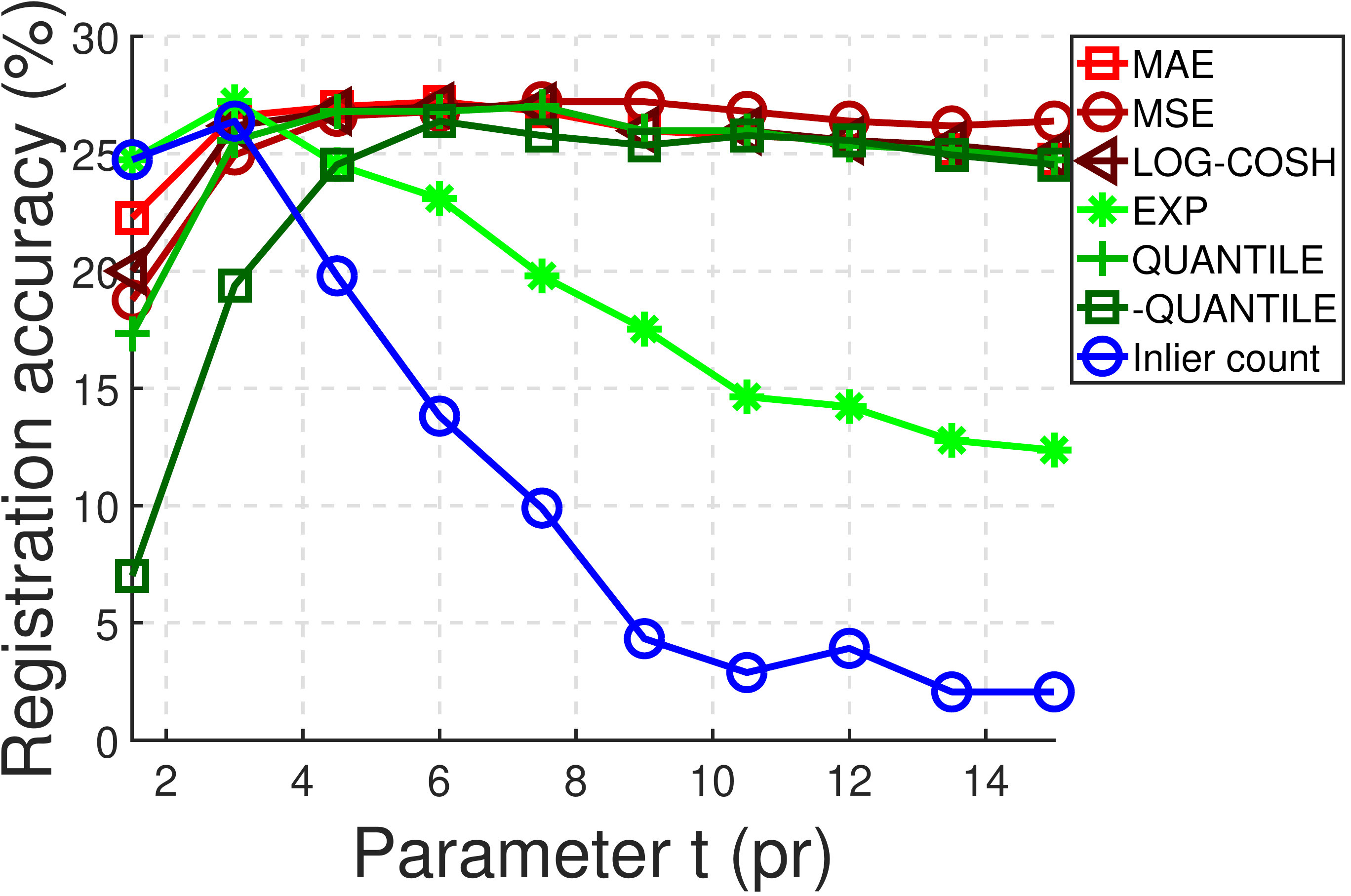}}
	\end{minipage}
	\begin{minipage}{0.246\linewidth}
		\centering
		\subfigure[$d_{rmse}$=2.5 pr]{
			\includegraphics[width=1\linewidth]{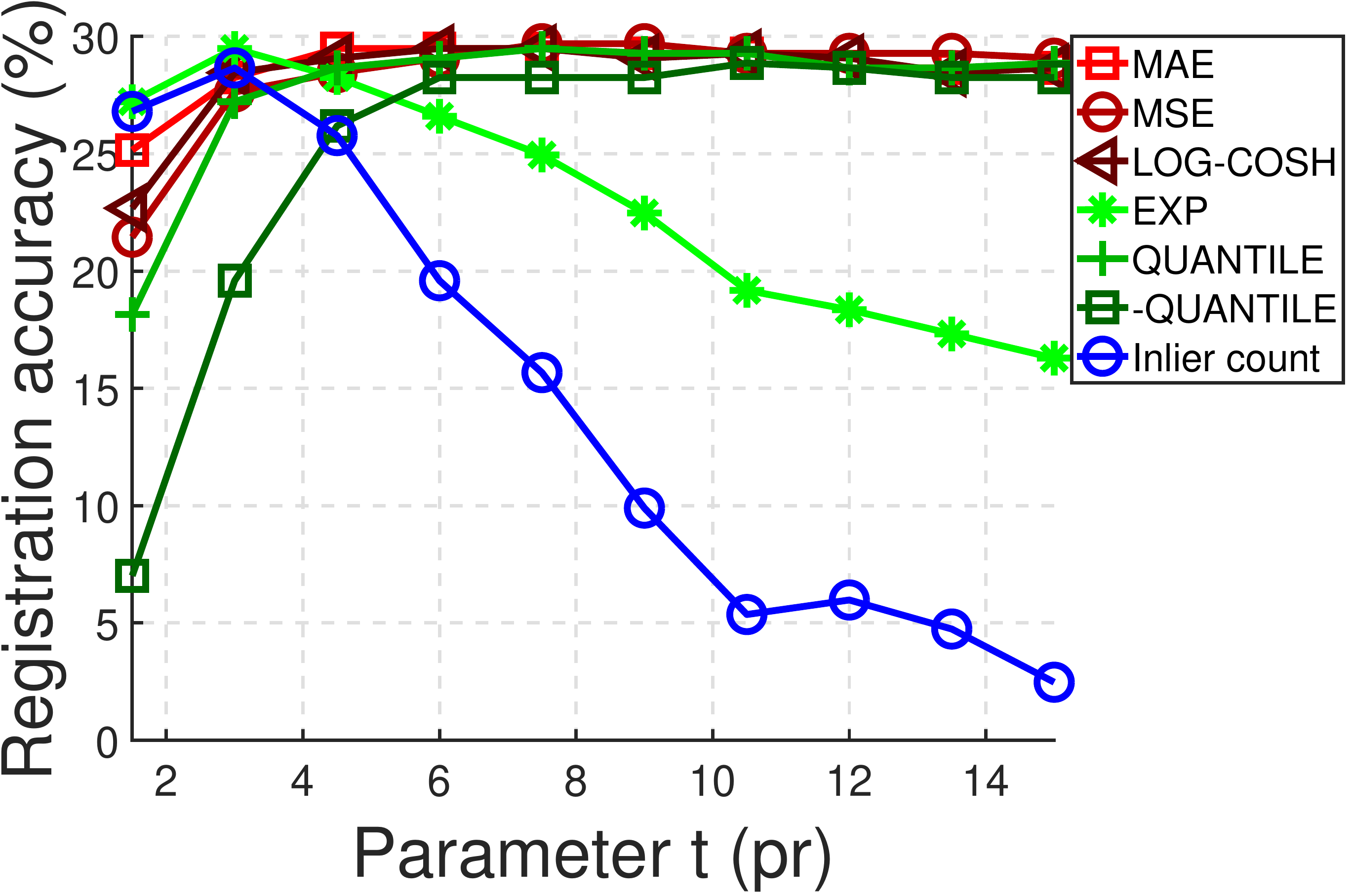}}
	\end{minipage}
	\begin{minipage}{0.246\linewidth}
		\centering
		\subfigure[$d_{rmse}$=3.0 pr]{
			\includegraphics[width=1\linewidth]{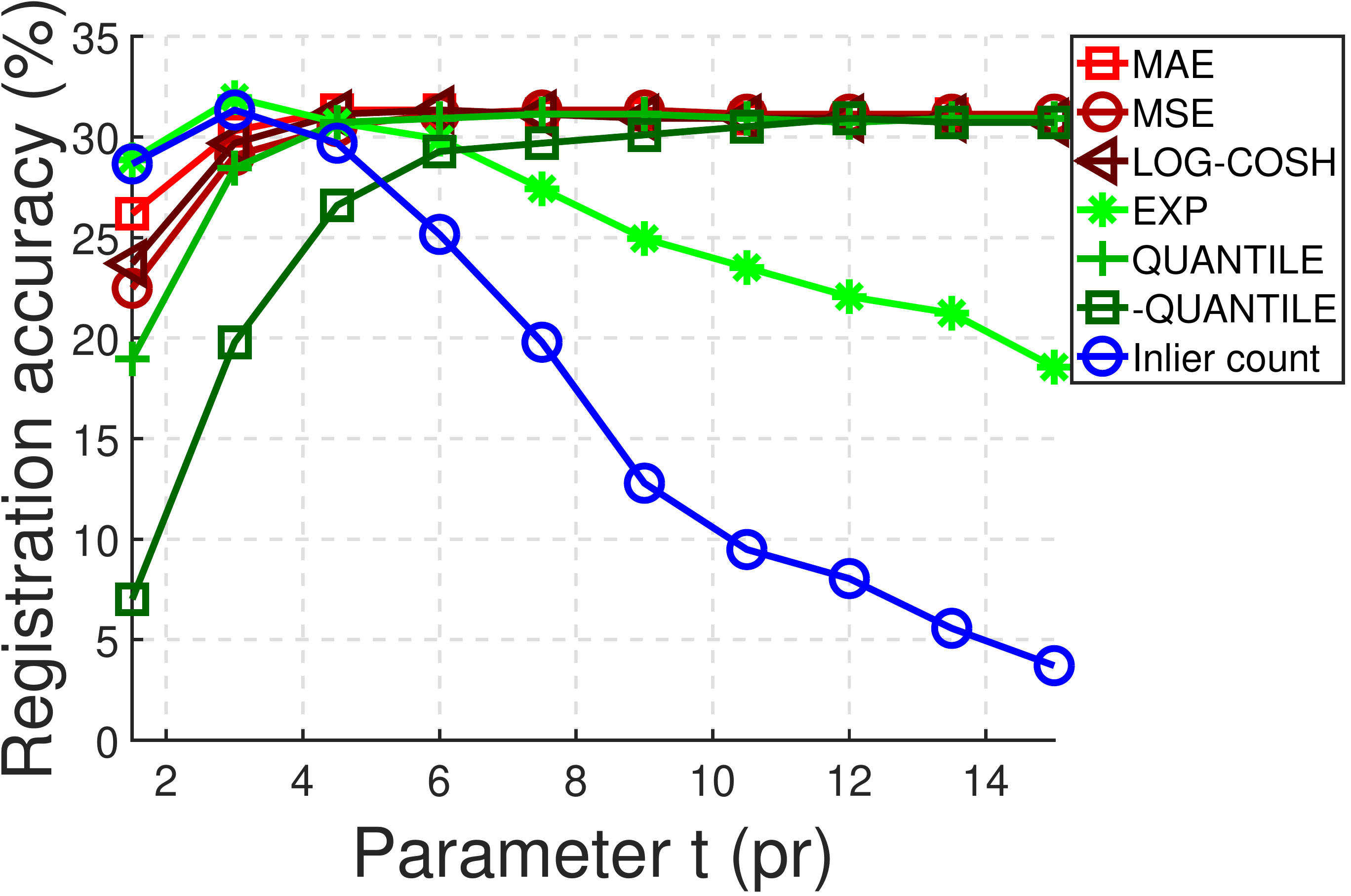}}
	\end{minipage}
	\begin{minipage}{0.246\linewidth}
		\centering
		\subfigure[$d_{rmse}$=3.5 pr]{
			\includegraphics[width=1\linewidth]{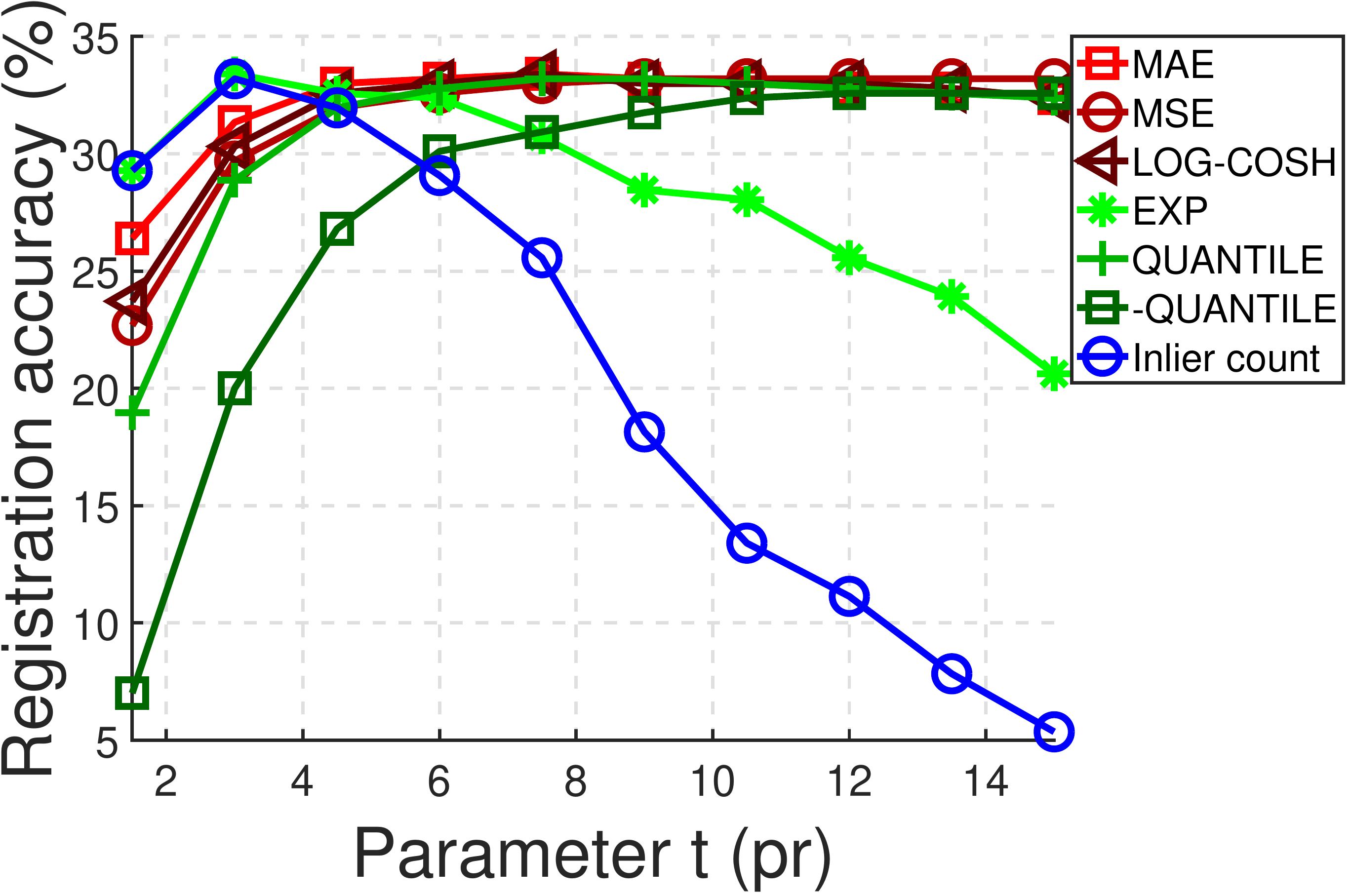}}
	\end{minipage}
	\begin{minipage}{0.246\linewidth}
		\centering
		\subfigure[$d_{rmse}$=4.0 pr]{
			\includegraphics[width=1\linewidth]{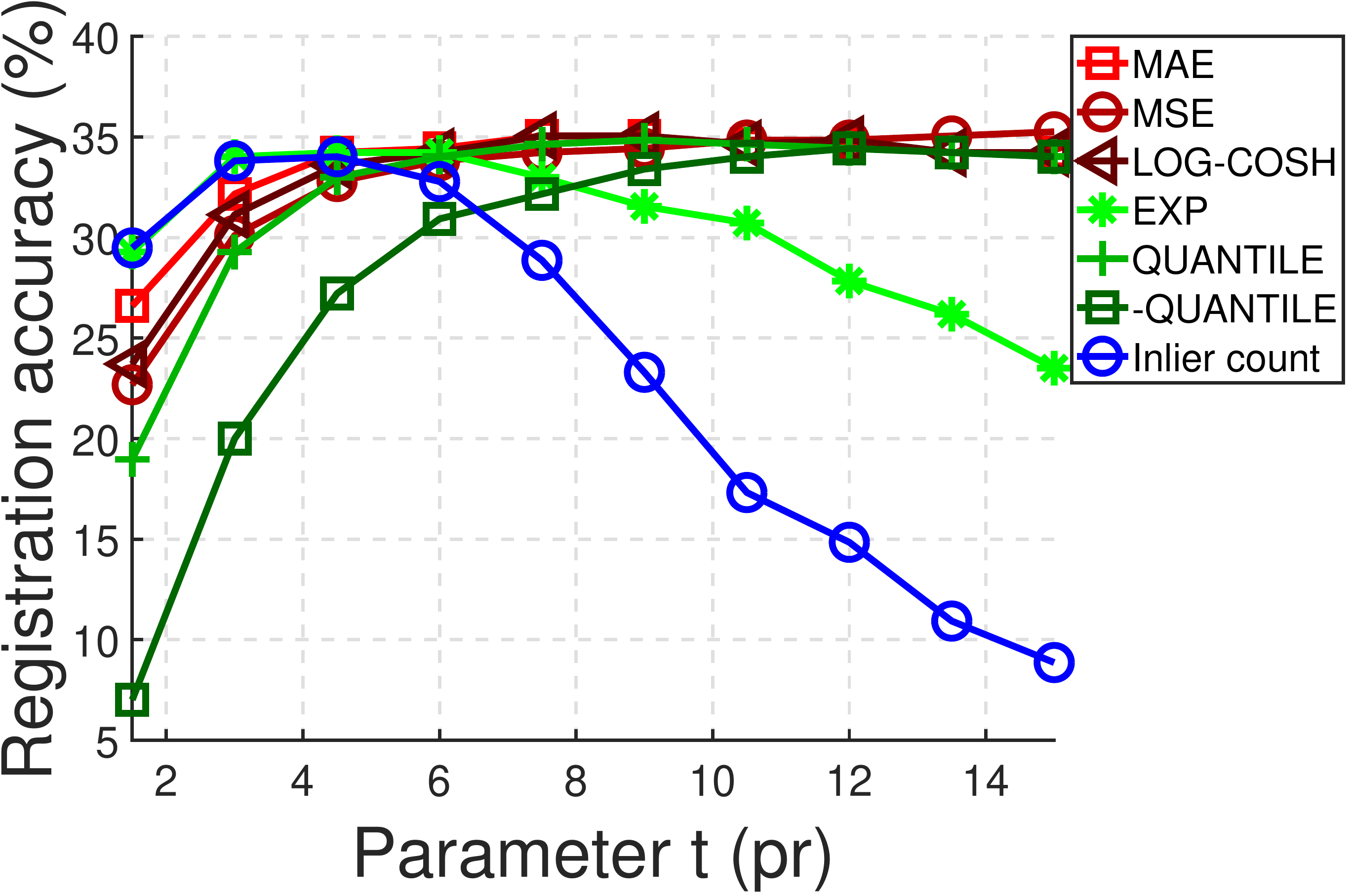}}
	\end{minipage}
	\caption{Sensitivity of tested metrics to parameter $t$ when varying RMSE thresholds.}
	\label{fig:para_r_RMSE}
\end{figure*}
\begin{figure*}[t]
	\begin{minipage}{0.246\linewidth}
		\centering
		\subfigure[400 iters.]{
			\includegraphics[width=1\linewidth]{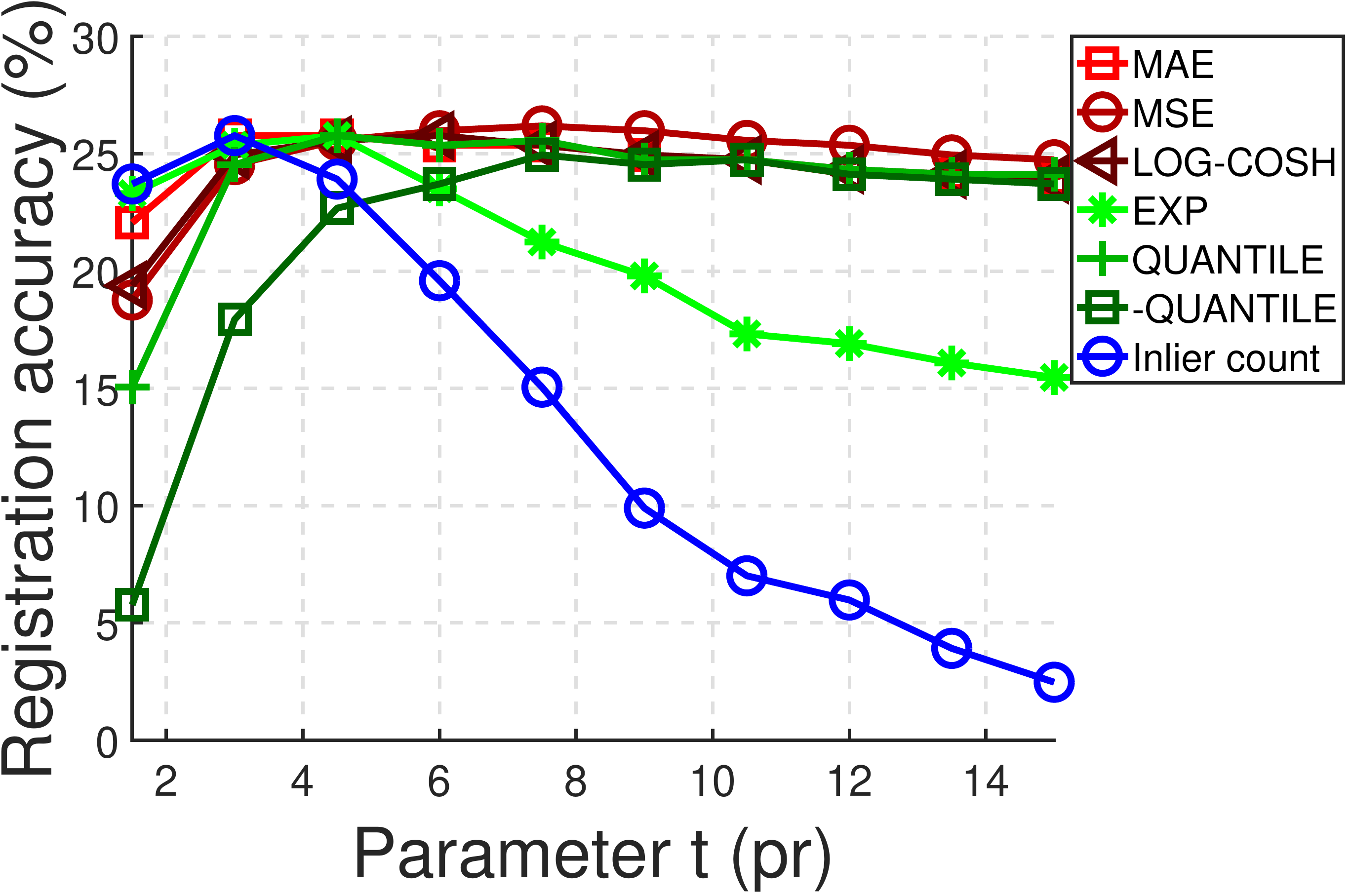}}
	\end{minipage}
	\begin{minipage}{0.246\linewidth}
		\centering
		\subfigure[600 iters.]{
			\includegraphics[width=1\linewidth]{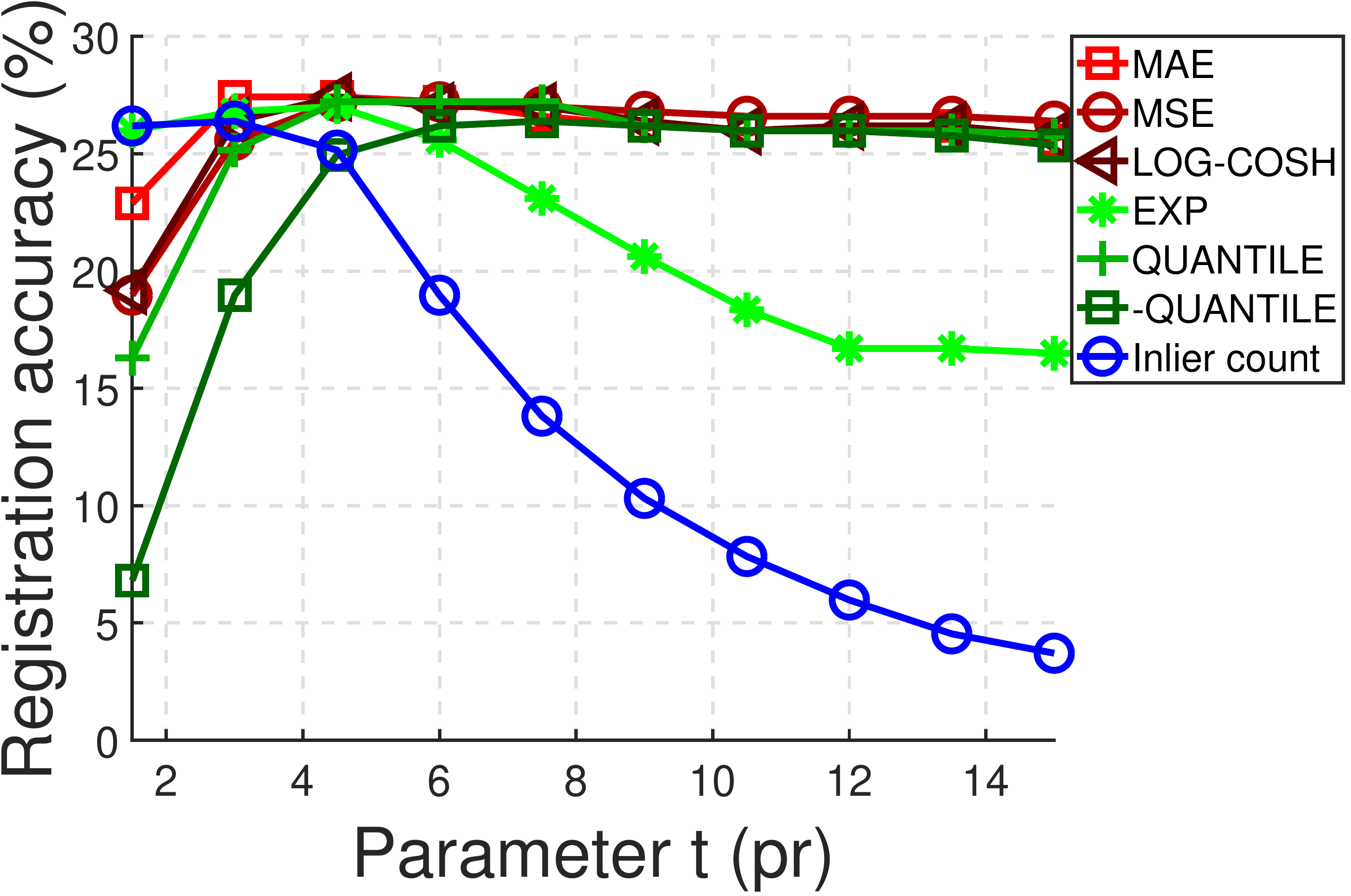}}
	\end{minipage}
	\begin{minipage}{0.246\linewidth}
		\centering
		\subfigure[800 iters.]{
			\includegraphics[width=1\linewidth]{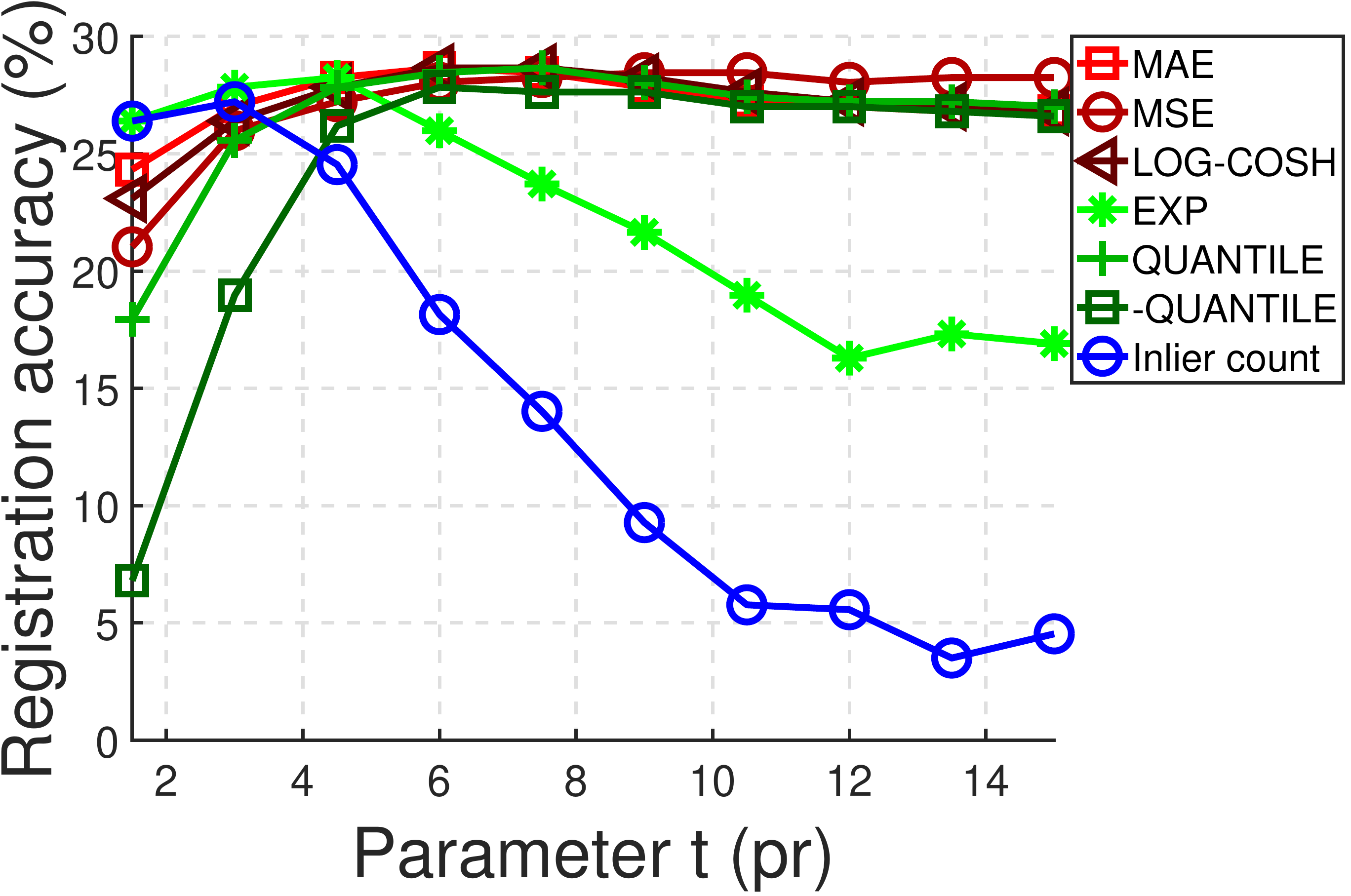}}
	\end{minipage}
	\begin{minipage}{0.246\linewidth}
		\centering
		\subfigure[1000 iters.]{
			\includegraphics[width=1\linewidth]{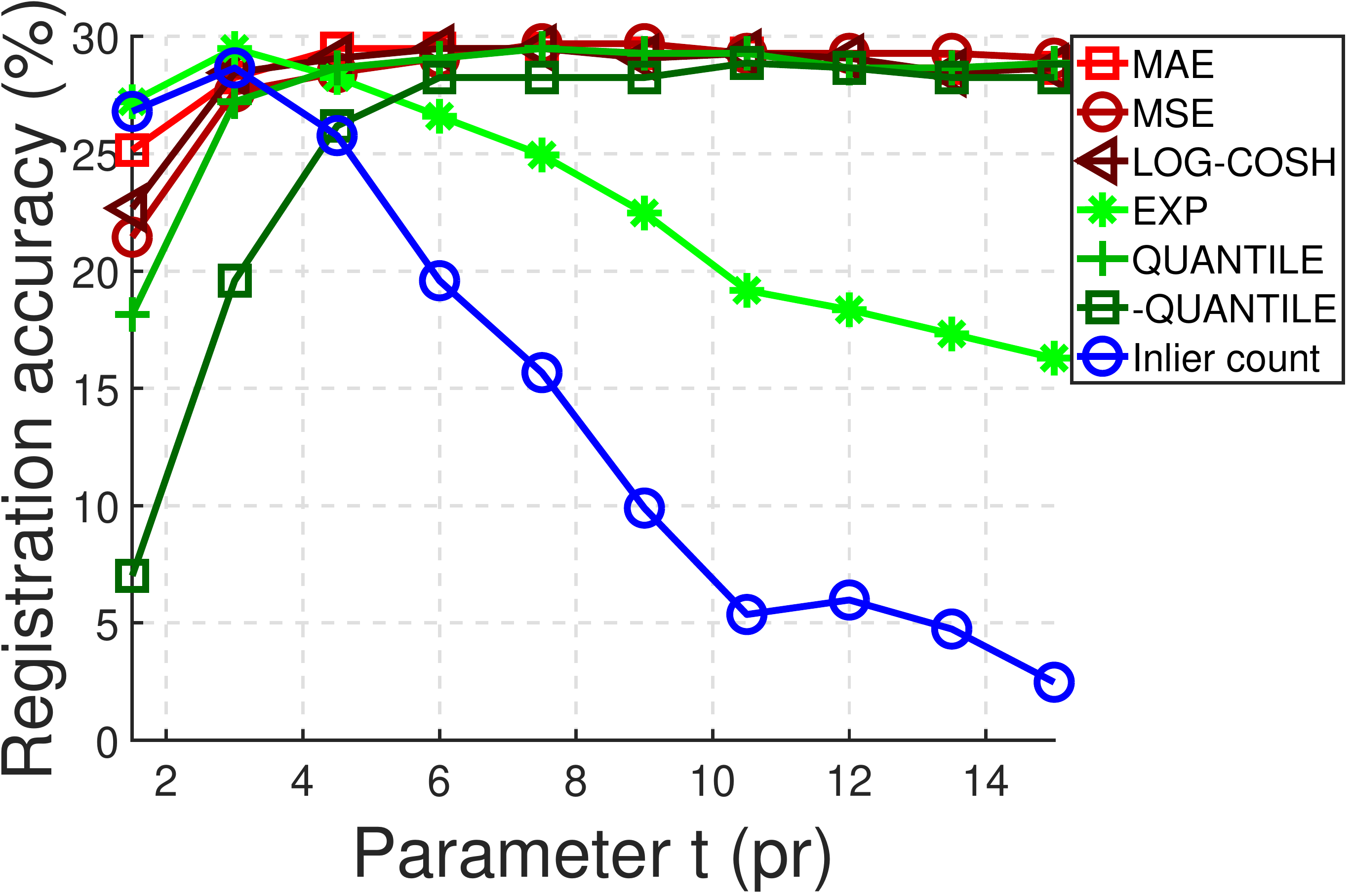}}
	\end{minipage}
	\caption{Sensitivity of tested metrics to parameter $t$ when varying RANSAC iterations.}
	\label{fig:para_r_iter}
\end{figure*}
All experiments were conducted in the point cloud library (PCL)~\cite{rusu20113d} with a 3.4 GHz processor and 16 GB RAM.
\begin{figure*}[t]
	\begin{minipage}{0.49\linewidth}
		\centering
		\subfigure[U3M]{
			\includegraphics[width=1\linewidth]{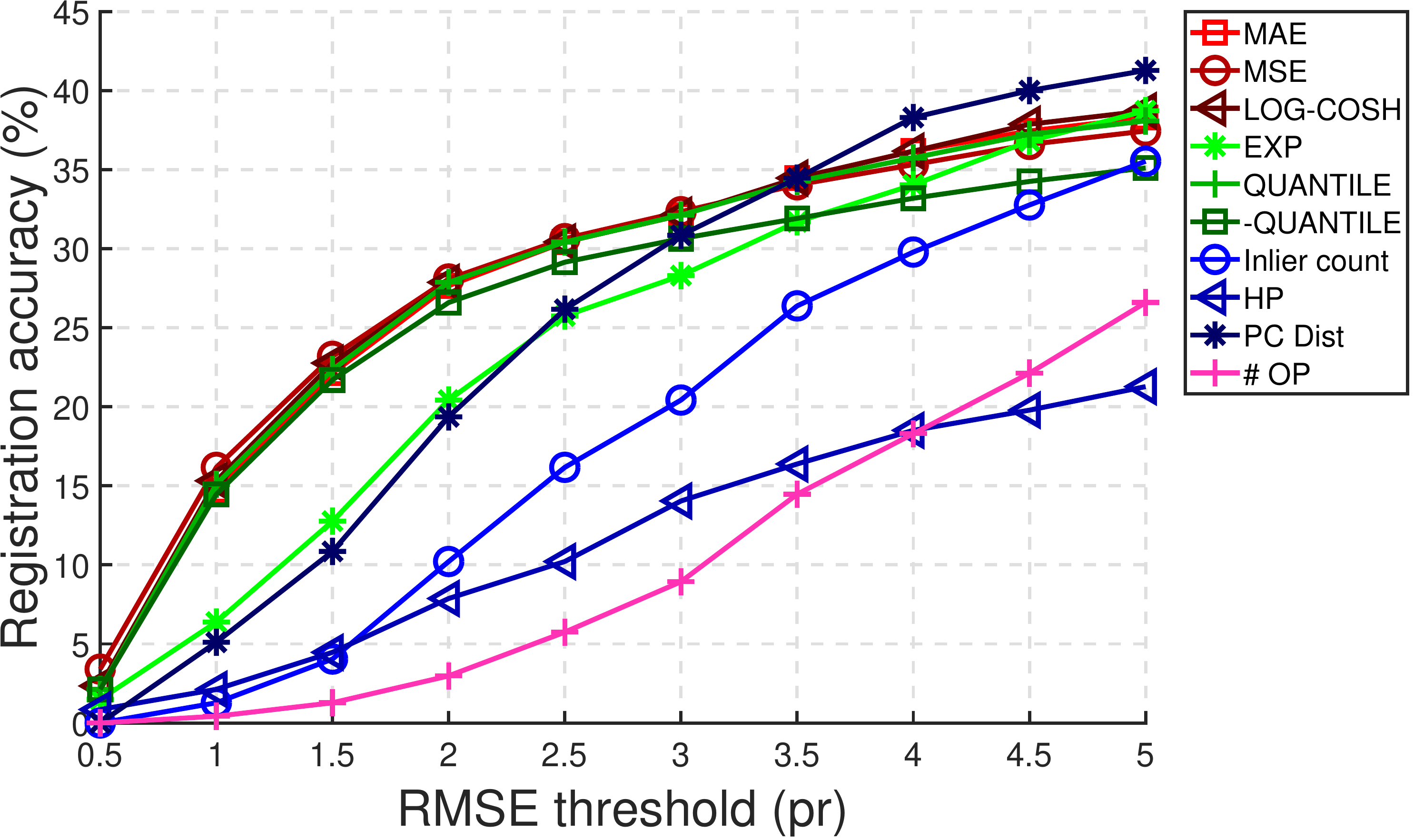}}
	\end{minipage}
	\begin{minipage}{0.49\linewidth}
		\centering
		\subfigure[BMR]{
			\includegraphics[width=1\linewidth]{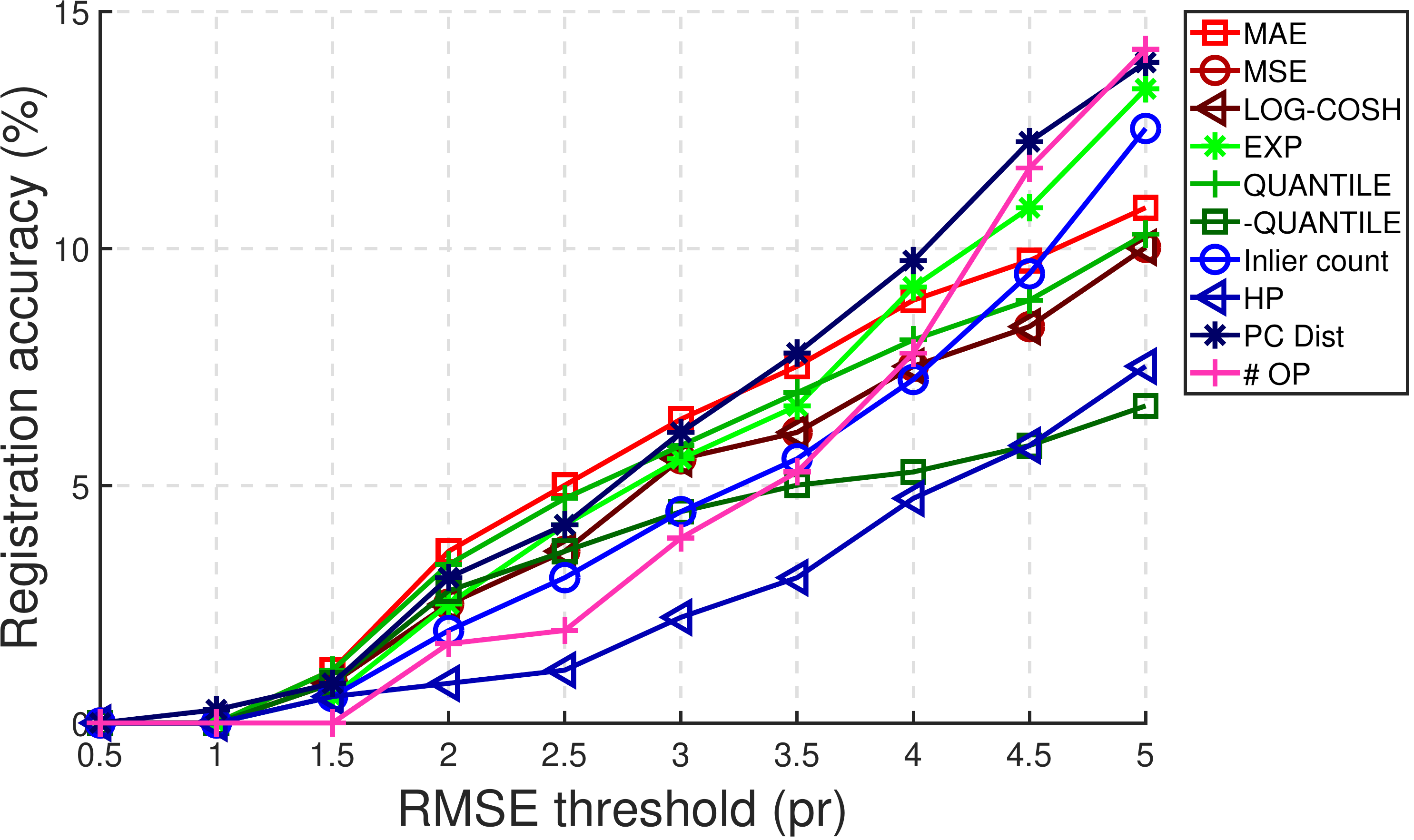}}
	\end{minipage}
	\begin{minipage}{0.49\linewidth}
		\centering
		\subfigure[U3OR]{
			\includegraphics[width=1\linewidth]{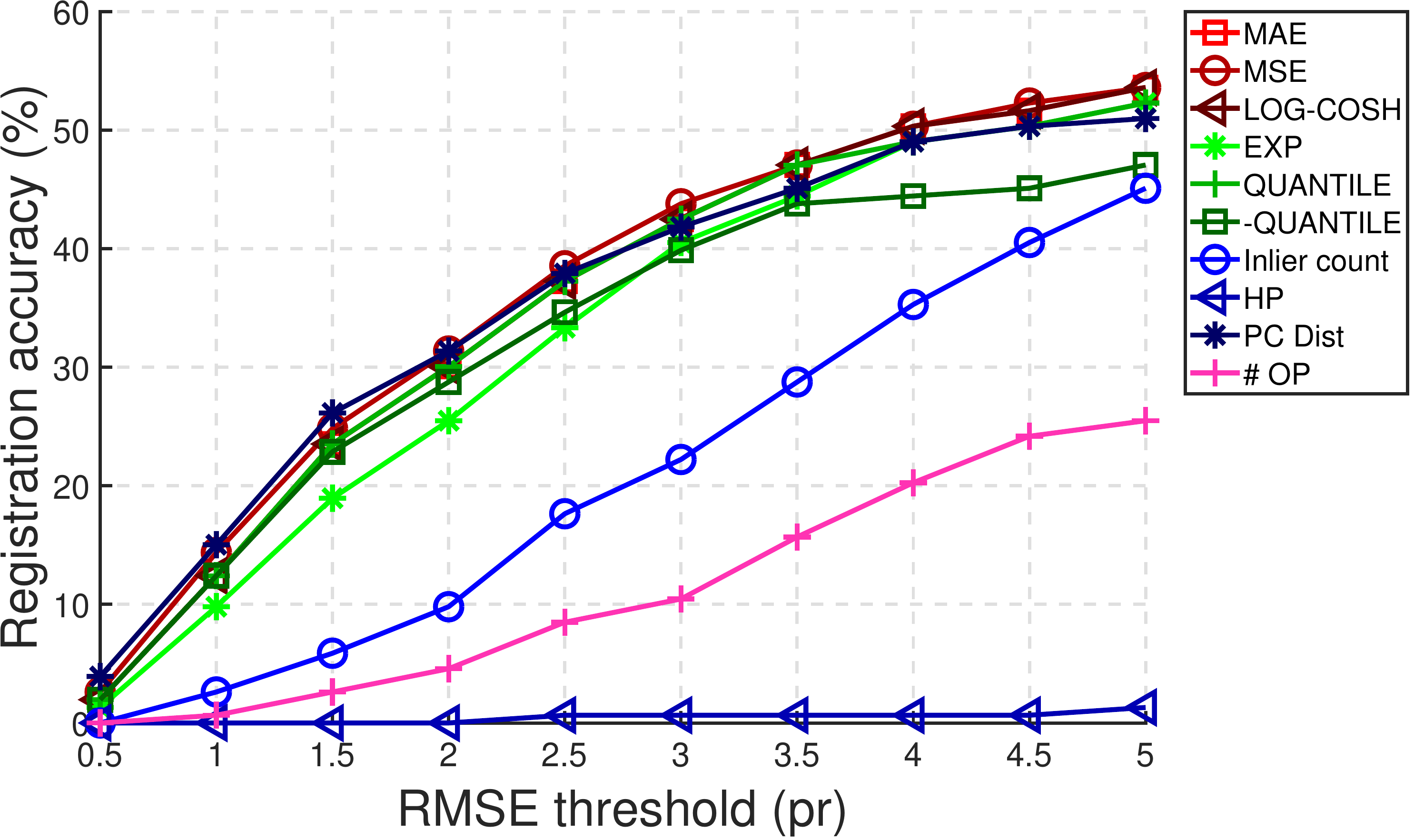}}
	\end{minipage}
	\begin{minipage}{0.49\linewidth}
		\centering
		\subfigure[BoD5]{
			\includegraphics[width=1\linewidth]{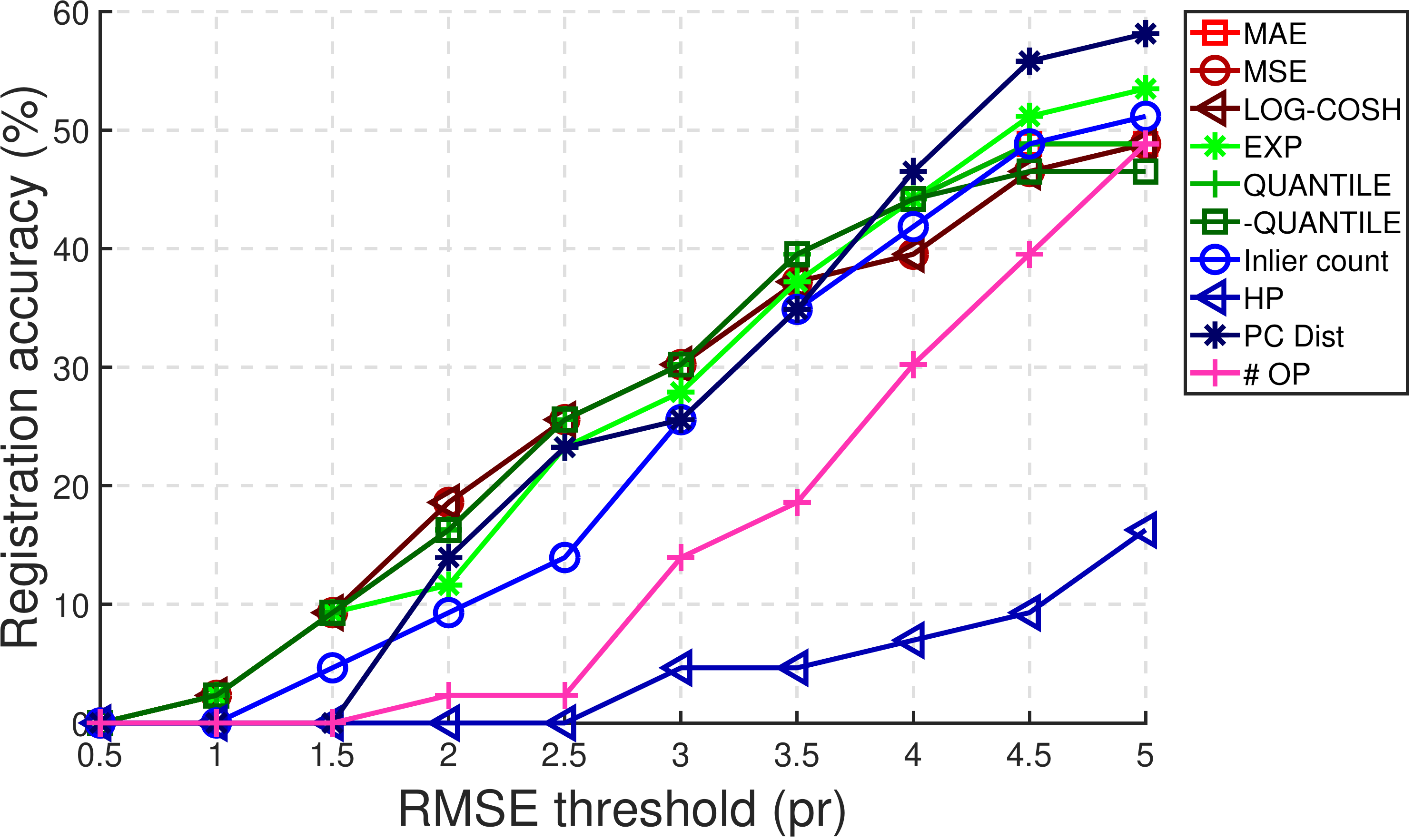}}
	\end{minipage}
	\caption{3D rigid registration accuracy performance of tested metrics on four experimental datasets with respect to different RMSE thresholds.}
	\label{fig:accuracy}
\end{figure*}

\subsection{Method Analysis Results}\label{subsec:anay_exp}
Method analysis experiments were conducted on the U3M dataset (expect for analyzing the parameter $t$), which is the largest-scale one among four considered datasets. Because the inlier count metric is the most commonly used one in RANSAC estimators and it is also a correspondences-based metric as the proposed ones, we will therefore treat it as a critical baseline in method analysis experiments. Note that other existing metrics will be thoroughly compared in comparative experiments (Sect.~\ref{subsec:comp_exp}).

In the following experiments, we set $d_{rmse}$ to 2.5 pr by default. The unit `pr' denotes point cloud resolution, i.e., the average of the distances from each point in a point cloud to the closest point to it.
\subsubsection{Varying the Number of Iterations}

We vary the number of iterations required by RANSAC with different hypothesis evaluation metrics. The results are shown in Fig.~\ref{fig:para_iter}.

One can see that the proposed metrics outperform the inlier count metric when varying the number of iterations from 200 to 2000. In addition, it is interesting to note that the proposed metrics generally achieve better performance with more iterations. This indicates that as more reasonable hypotheses are generated, our metrics can reliably filter out them. By contrast, the inlier count metric often misses them.
\subsubsection{Varying RANSAC Estimators}
\begin{table}[t]
	\centering
	\caption{Registration accuracy results of different RANSAC estimators with different hypothesis evaluation metrics.}
	\label{tab:vary_est}
	\scalebox{1}{
		\begin{tabular}{|c| c|c| c|}
			\hline
			&\bf RANSAC~\cite{fischler1981random}&\bf 2SAC-GC~\cite{yang2017multi}&\bf 1P-RANSAC~\cite{guo2015integrated}\\
			\hline
			MAE&0.2948  &0.3608&0.2742\\
			\hline
			MSE&\bf 0.2969&0.3608&\bf0.2784\\
			\hline
			LOG-COSH& 0.2948&\bf0.3629&0.2722\\
			\hline
			EXP&0.2948&0.3031&0.2392\\
			\hline
			QUANTILE&0.2495&0.3608&0.2722\\
			\hline
			-QUANTILE&0.2825&0.3216&0.2433\\
			\hline
			Inlier count&0.1567&0.1897&0.1835\\
			\hline
	\end{tabular}}
\end{table}

A good metric should be general. We specifically check the effectiveness of our metrics when using different RANSAC estimators including RANSAC~\cite{fischler1981random}, 2-point sample consensus with global constraints (2SAC-GC)~\cite{yang2017multi}, and 1-point RANSAC (1P-RANSAC)~\cite{guo2015integrated}. The results are reported in Table~\ref{tab:vary_est}. 

Clearly, the performance boosting is consistent when replacing the inlier count metric by our ones for all tested RANSAC estimators. It suggests that our metrics are general to different RANSAC estimators.
\subsubsection{Sensitivity to Parameter $t$}
For correspondences-based metrics, they will face the problem of tuning  parameter $t$ (Eq.~\ref{eq:para_t}). By default, we set $t$ to 7.5 pr because the support radius of a local descriptor is suggested to 15 pr~\cite{guo2013rotational,tombari2010unique,yang2017RCS_jrnl} and the inlier judgment threshold for correspondences is suggested to be half of the support radius~\cite{guo2016comprehensive,yang2017toldi}. However, our experiments will show that the optimum value of $t$ usually varies with different application scenarios and data qualities. Moreover, finding the optimum $t$ requires tedious and very careful parameter tuning work. {\textit{ In other words, a good metric for RANSAC hypotheses should be robust the parameter $t$.}} We have conducted a set of experiments as following to demonstrate that our metrics are ultra robust to parameter $t$ (varying from 1 pr to 15 pr), which making them good gifts to practical applications.

Specifically, we examine the sensitivity of metrics to $t$ when changing datasets (i.e., varying application scenarios and data modalities), RMSE thresholds (i.e., varying accuracy requirements for registration results), and RANSAC iterations (i.e., varying other RANSAC parameters).
\\\\\noindent\textbf{Varying datasets} As shown in Fig.~\ref{fig:para_r_dataset}, it is clear that the proposed metrics (expect for EXP) achieve quit stable performance on all datasets if $t$ is greater than 4 pr. By contrast, the performance of the inlier count metric fluctuates dramatically on all datasets. Moreover, the optimum value of $t$ to the inlier count metric varies with datasets, indicating its sensitivity to data modality and application scenario changes.
\\\\\noindent\textbf{Varying RMSE thresholds}
As witnessed by Fig.~\ref{fig:para_r_RMSE}, we find that the proposed metrics neatly outperform the inlier count metric with small $d_{rmse}$ values. This suggests that our metrics can produce very accurate registrations. Although by carefully tuning the parameter $t$, the inlier count metric can achieve comparable performance as ours when $d_{rmse}$ is greater than 2.0 pr, the optimum value of $t$ to our metrics can general to all tested cases which have different accuracy requirements. 
\\\\\noindent\textbf{Varying RANSAC iterations}
As shown in Fig.~\ref{fig:para_r_iter}, we can see that the inlier count metric is still very susceptible to parameter $t$ when varying RANSAC iterations. By contrast, most of the proposed metrics, e.g., MAE and MSE, are robust to varying values of $t$ under different RANSAC iterations.

All results in Fig.~\ref{fig:para_r_dataset}, Fig.~\ref{fig:para_r_RMSE}, and Fig.~\ref{fig:para_r_iter} demonstrate that most of our proposed metrics hold strong robustness to parameter $t$. In other words, {\textit{one can use a fixed value of $t$ for our metrics to accomplish accurate registrations in scenarios with different application contexts, data modalities, and different accuracy demands.}} This is critical to practical applications.

\subsection{Comparative Results}\label{subsec:comp_exp}
Here, we compare the proposed metrics with {\textit{all known  metrics to us}} for RANSAC hypotheses in the context of 3D rigid registration. They include the inlier count, Huber penalty (HP)~\cite{rusu2009fast}, point cloud distance (PC Dist)~\cite{yang2016fast}, and the number of overlapped points (\# OP)~\cite{Quan2018Local}. We set $t$ to 7.5 pr for our metrics based on experiments in Sect.~\ref{subsec:anay_exp} while keeping parameters of compared metrics identical to the suggested settings in their original publications. Note that all metrics are tested in the same RANSAC pipeline to ensure a fair comparison.
\subsubsection{Results on Different Datasets}
\begin{figure}[t]
	\centering
	\includegraphics[width=0.8\linewidth]{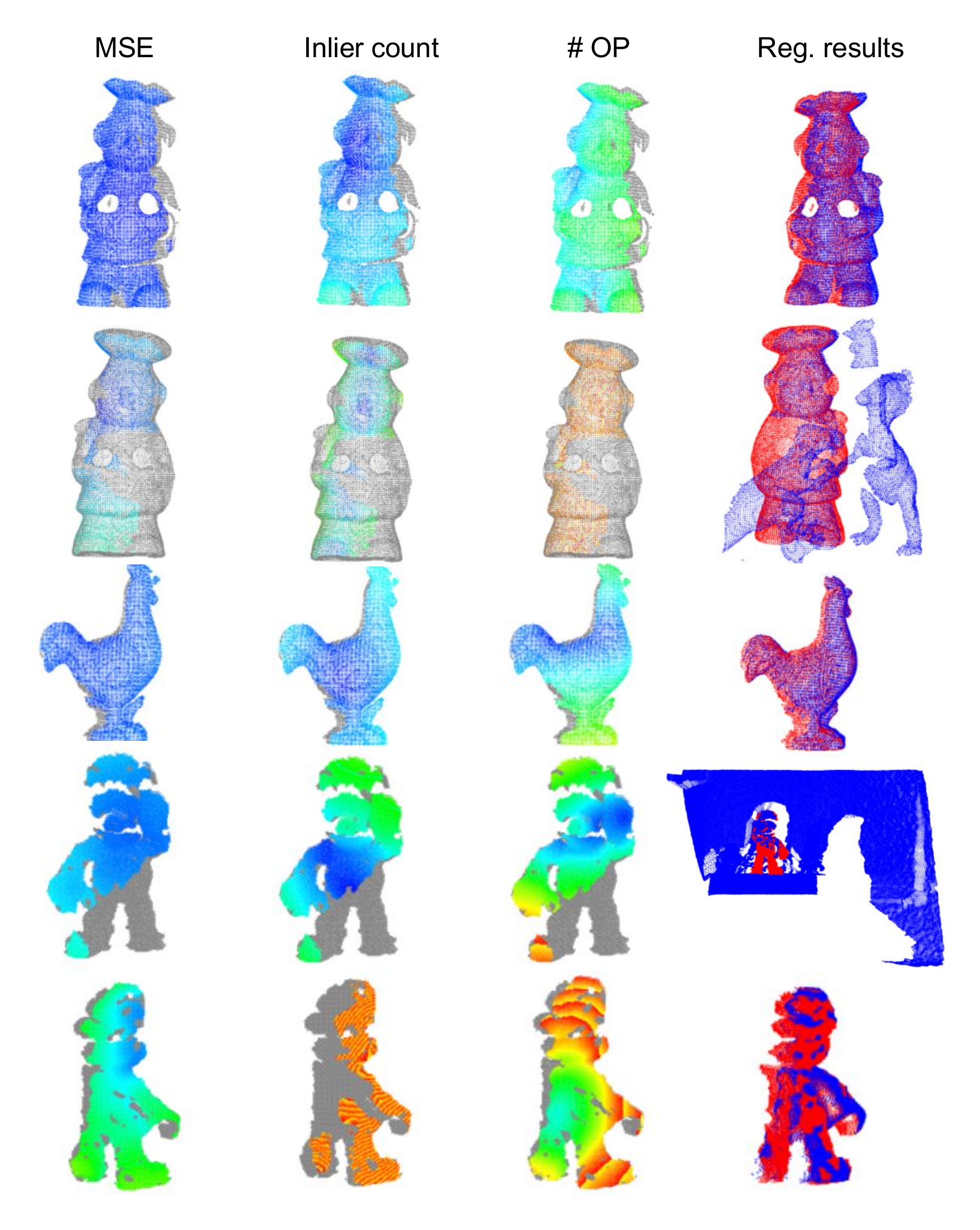}\\
	\caption{Visualization of point-wise registration errors by (from left to right) MSE, inlier count, and \# OP metrics and the registration results by MSE. The first column to the third one: point-wise registration errors are rendered by pseudo-color (blue $\rightarrow$ red: small errors to large errors).}
	\label{fig:visualization}
\end{figure}

In this experiment, we vary $d_{rmse}$ from 0.5 pr to 5.0 pr with a step of 0.5 pr. The results are presented in Fig.~\ref{fig:accuracy}. Several observations can be made from the figure.
\begin{figure}[t]
	\begin{minipage}{0.49\linewidth}
		\centering
		\subfigure[]{
			\includegraphics[width=1\linewidth]{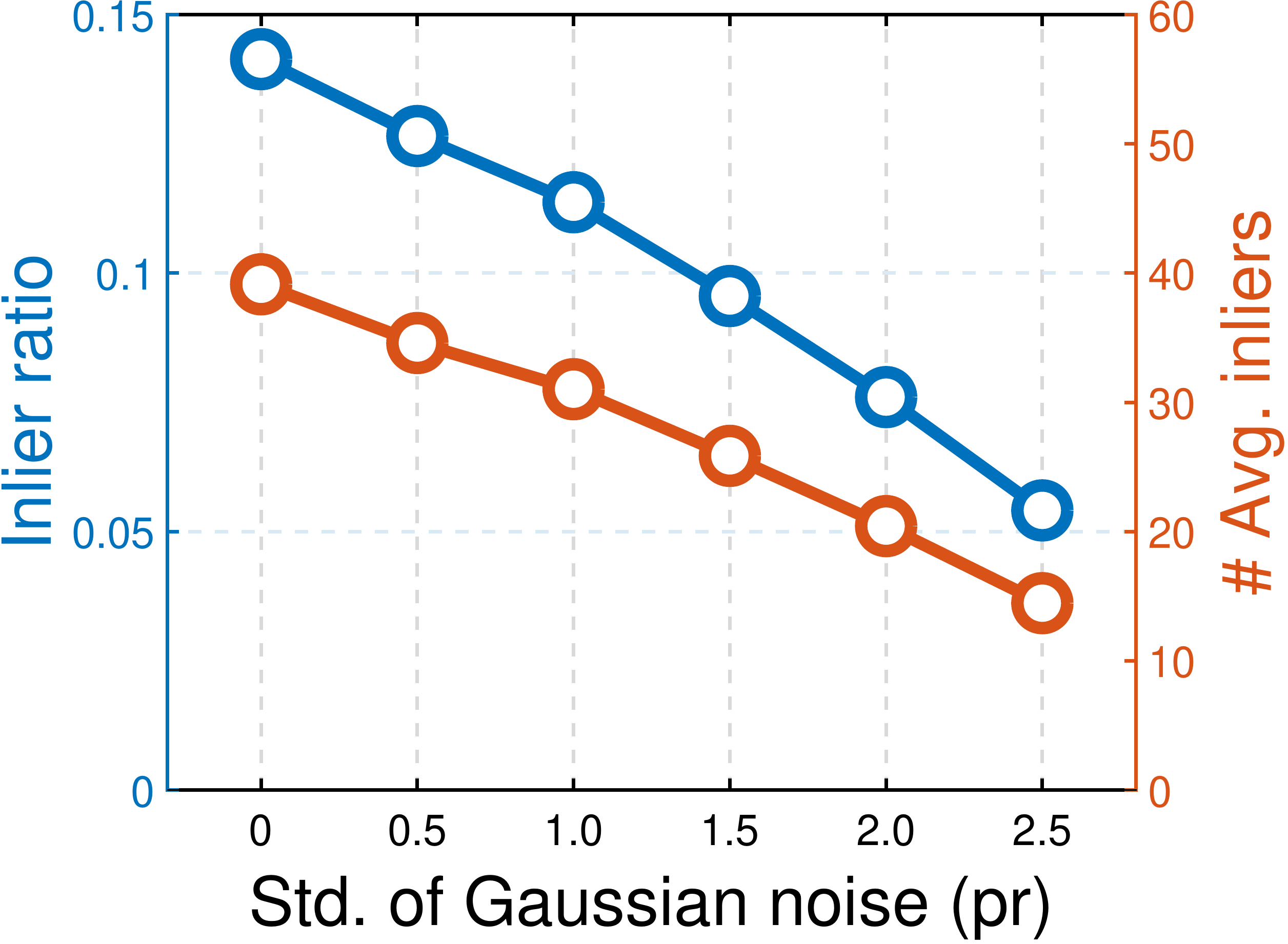}}
	\end{minipage}
	\begin{minipage}{0.49\linewidth}
		\centering
		\subfigure[]{
			\includegraphics[width=1\linewidth]{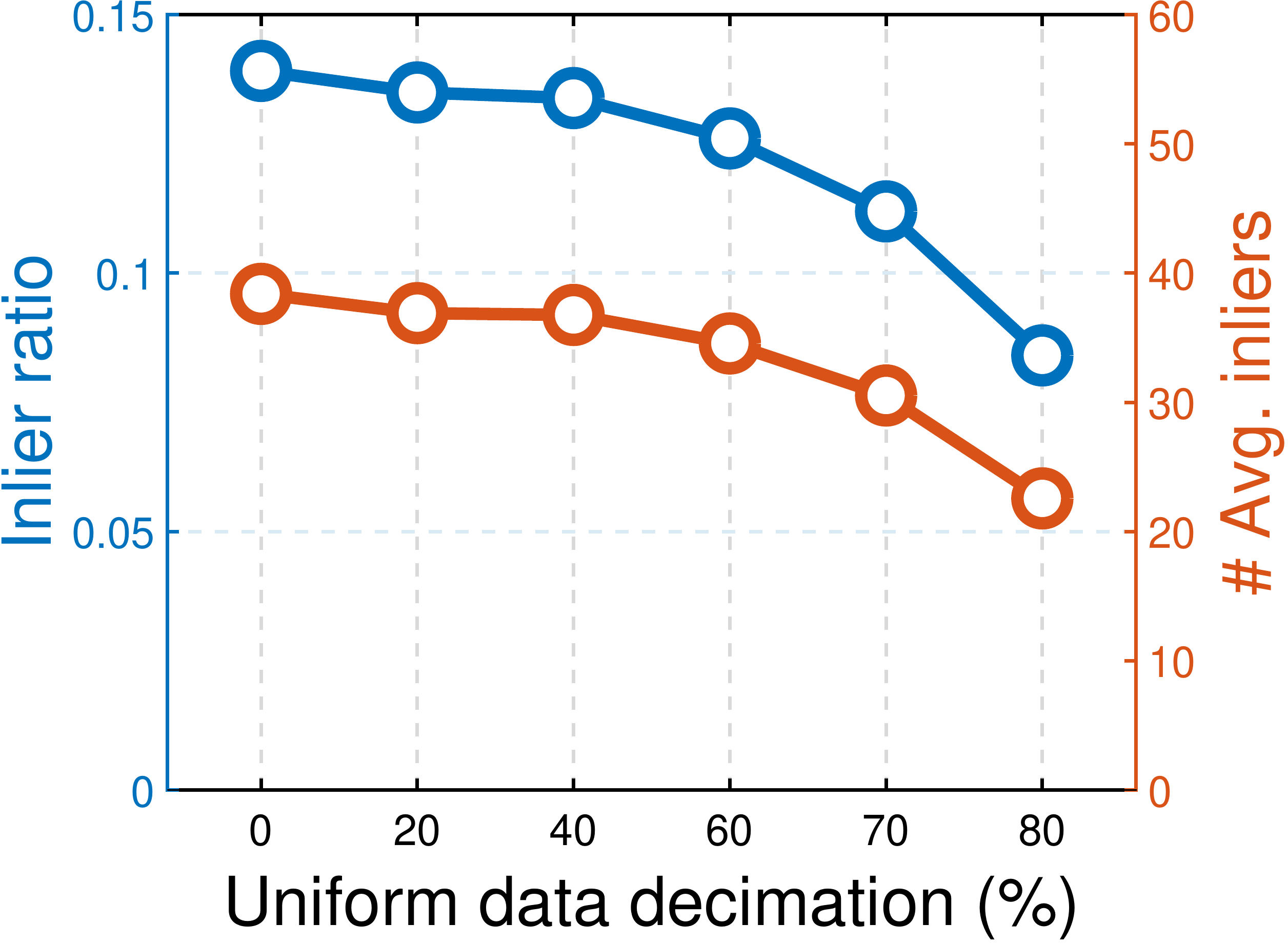}}
	\end{minipage}
	\begin{minipage}{0.49\linewidth}
		\centering
		\subfigure[]{
			\includegraphics[width=1\linewidth]{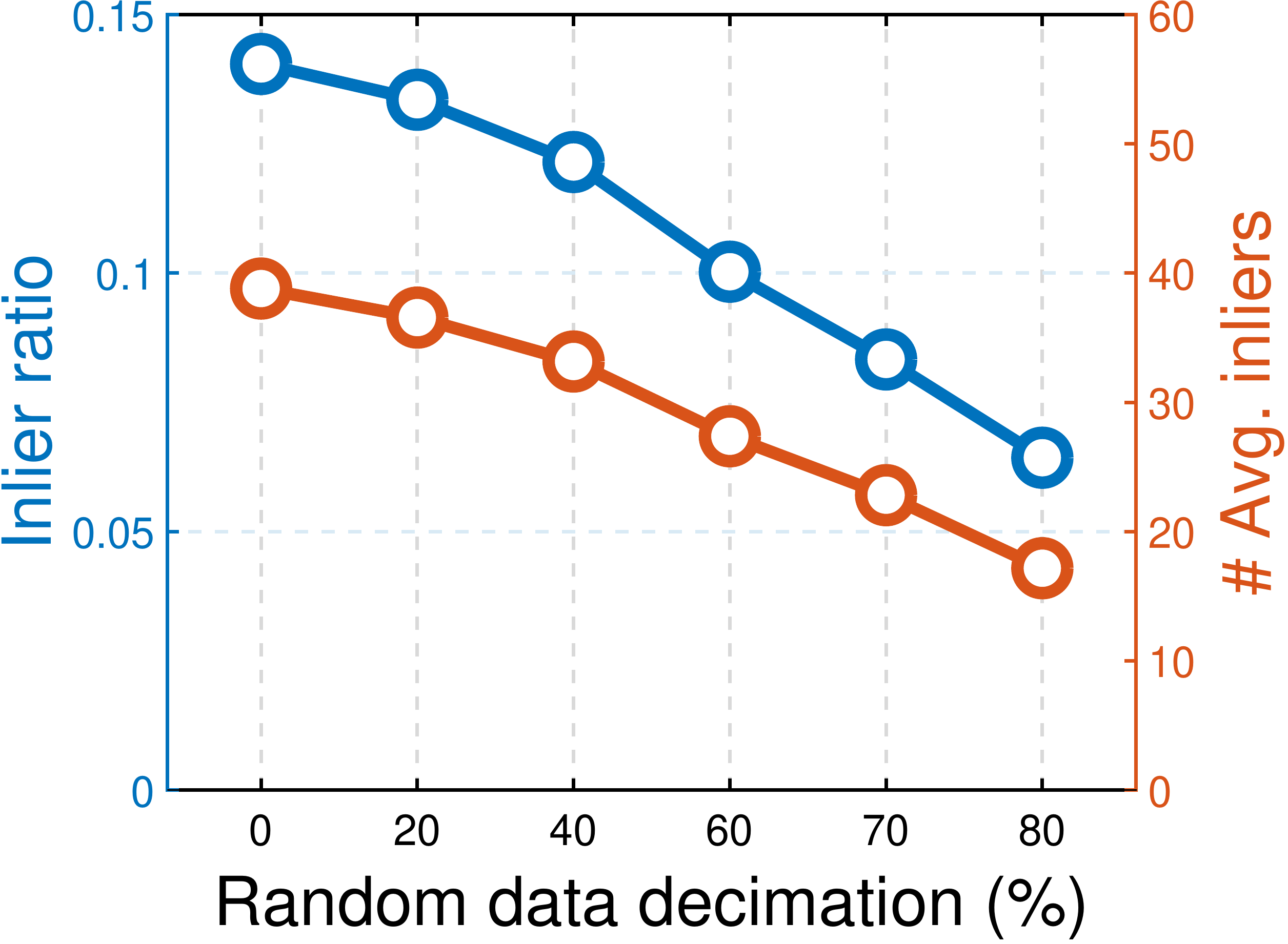}}
	\end{minipage}
	\begin{minipage}{0.49\linewidth}
		\centering
		\subfigure[]{
			\includegraphics[width=1\linewidth]{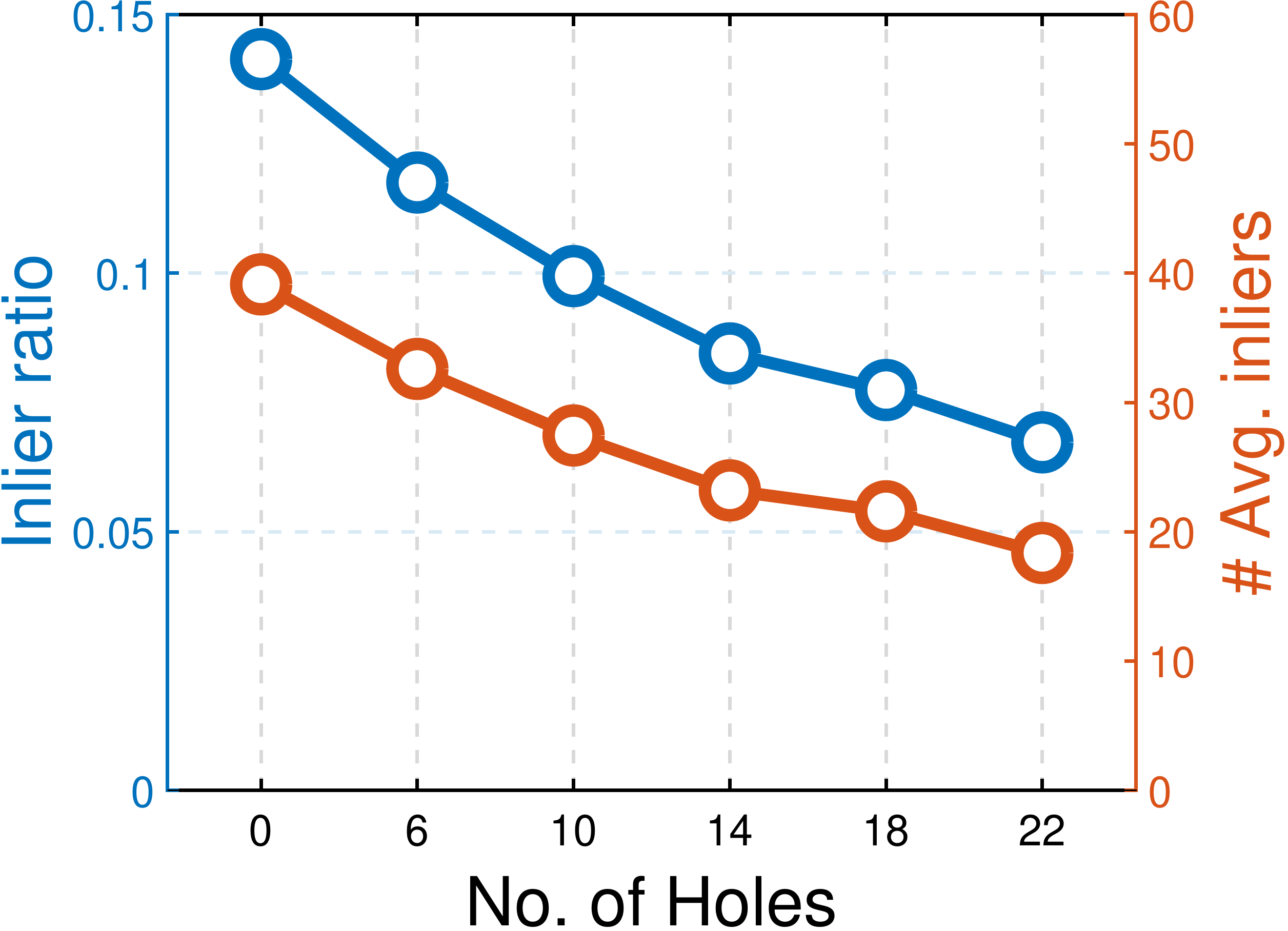}}
	\end{minipage}
	\begin{minipage}{0.49\linewidth}
		\centering
		\subfigure[]{
			\includegraphics[width=1\linewidth]{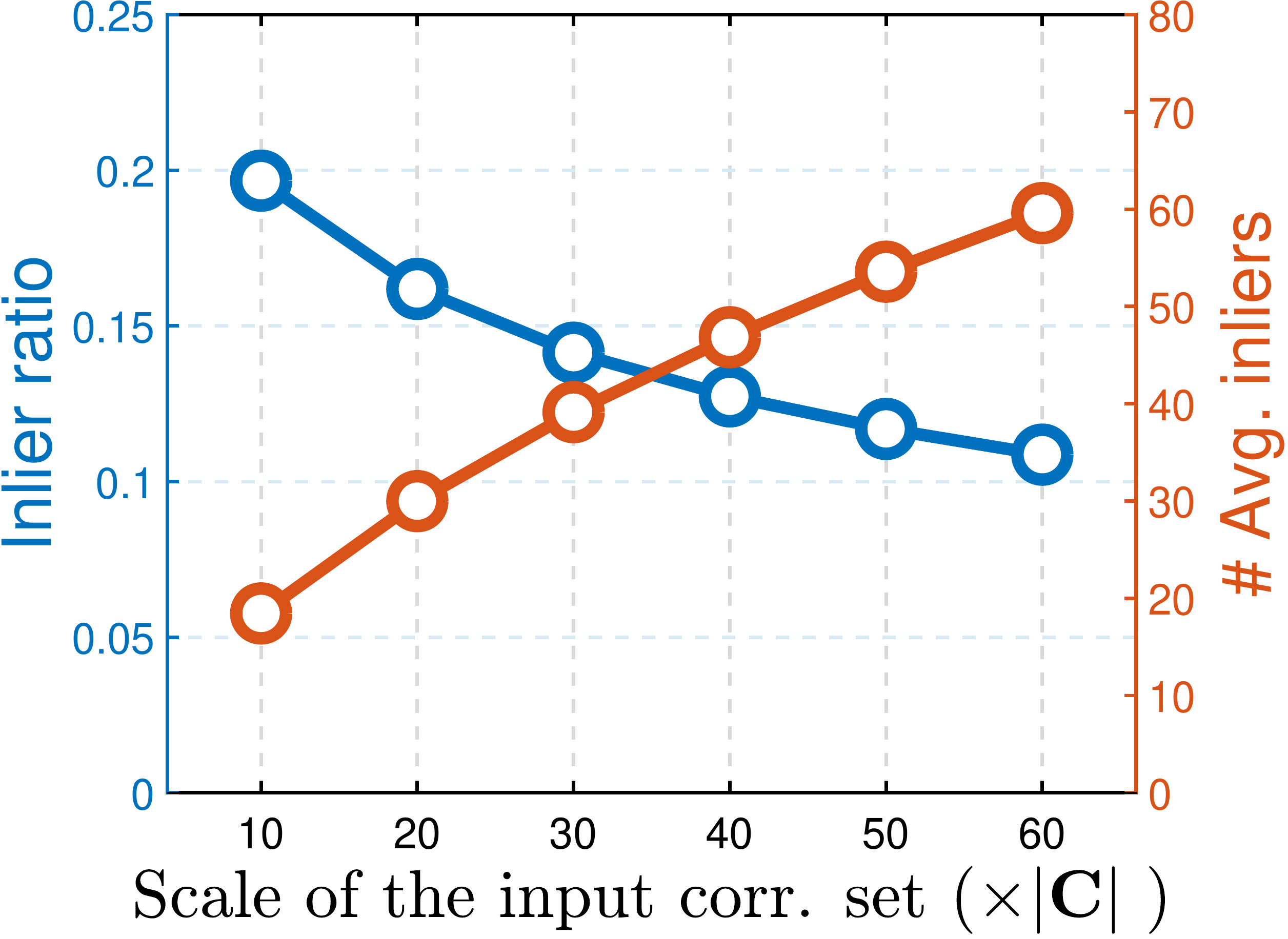}}
	\end{minipage}
	\begin{minipage}{0.49\linewidth}
		\centering
		\subfigure[]{
			\includegraphics[width=1\linewidth]{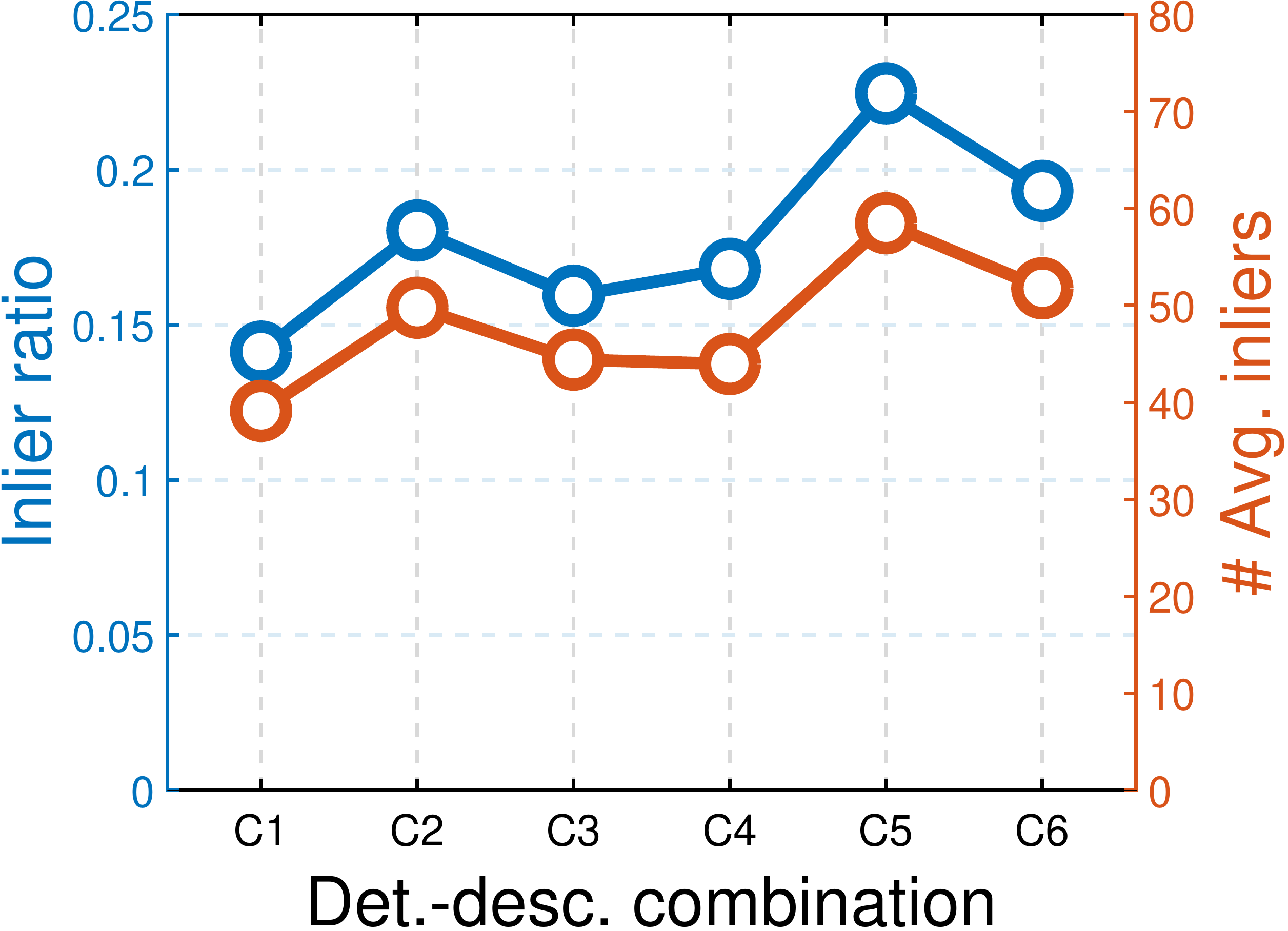}}
	\end{minipage}
	\caption{Information in terms of the inlier ratio and the number of inliers of the input correspondence sets under different experimental configurations.}
	\label{fig:nuisance_info}
\end{figure}

\begin{figure*}[t]
	\begin{minipage}{0.49\linewidth}
		\raggedleft
		\subfigure[]{
			\includegraphics[width=0.9\linewidth]{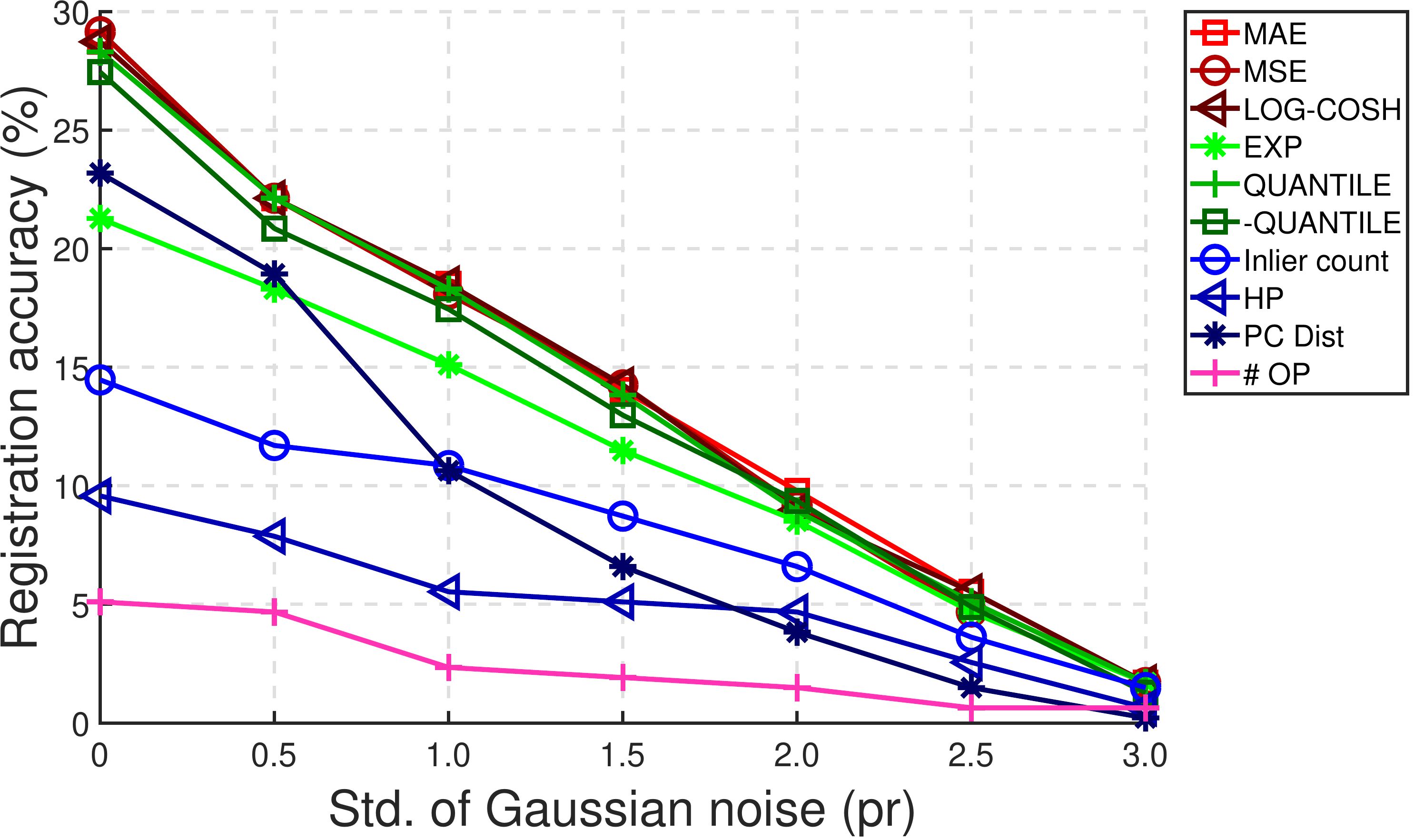}}
	\end{minipage}
	\begin{minipage}{0.49\linewidth}
		\raggedright
		\subfigure[]{
			\includegraphics[width=0.9\linewidth]{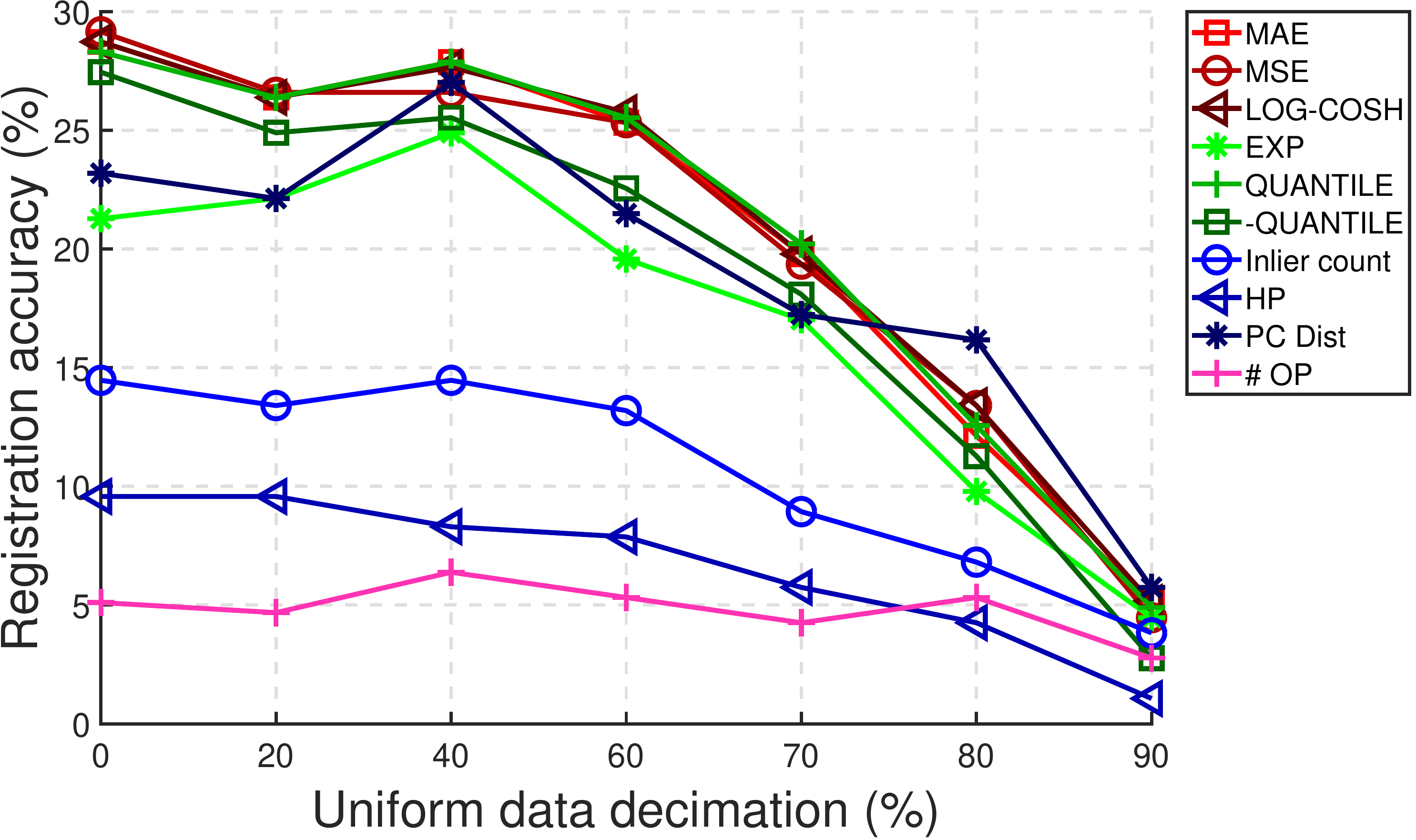}}
	\end{minipage}
	\begin{minipage}{0.49\linewidth}
		\raggedleft
		\subfigure[]{
			\includegraphics[width=0.9\linewidth]{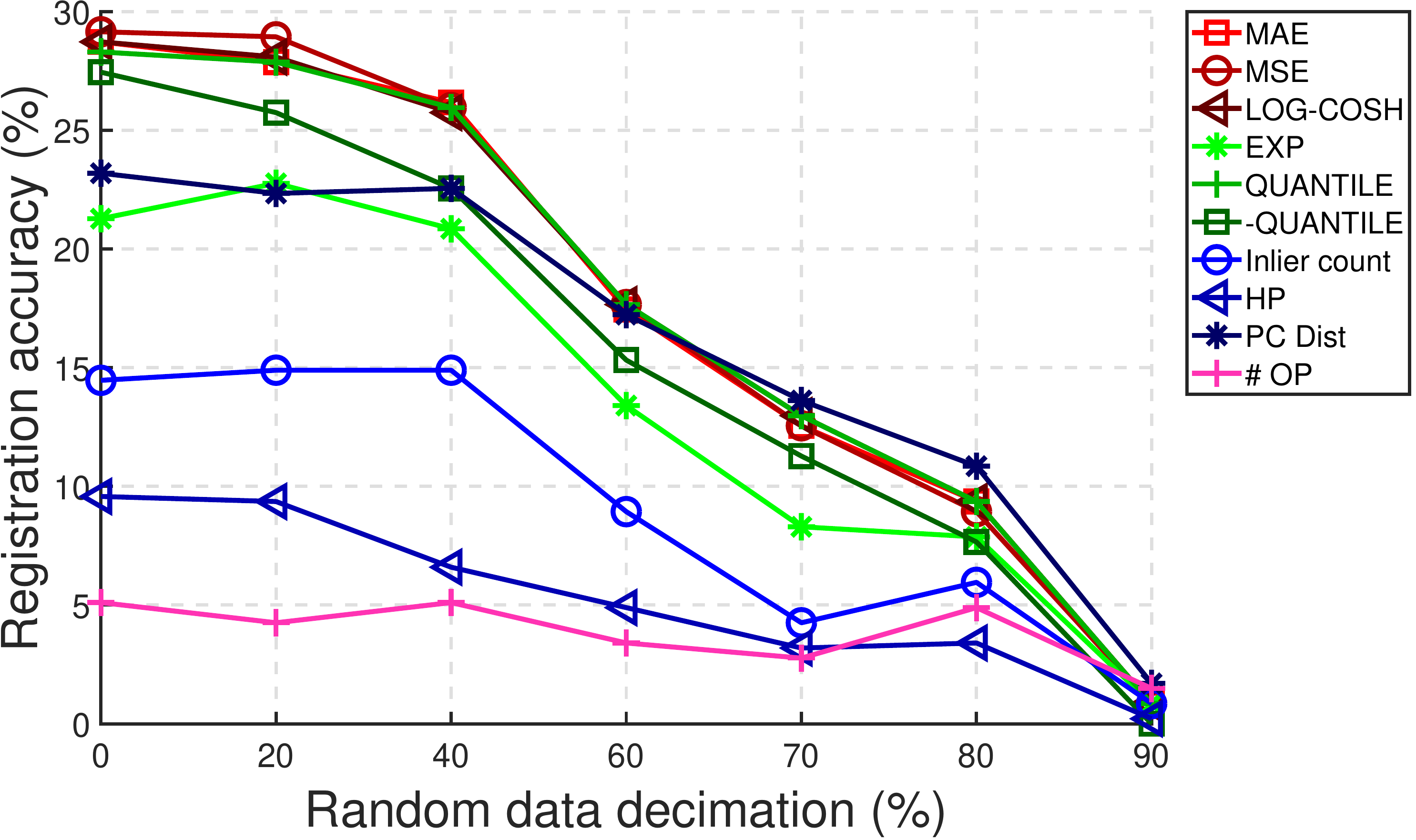}}
	\end{minipage}
	\begin{minipage}{0.49\linewidth}
		\raggedright
		\subfigure[]{
			\includegraphics[width=0.9\linewidth]{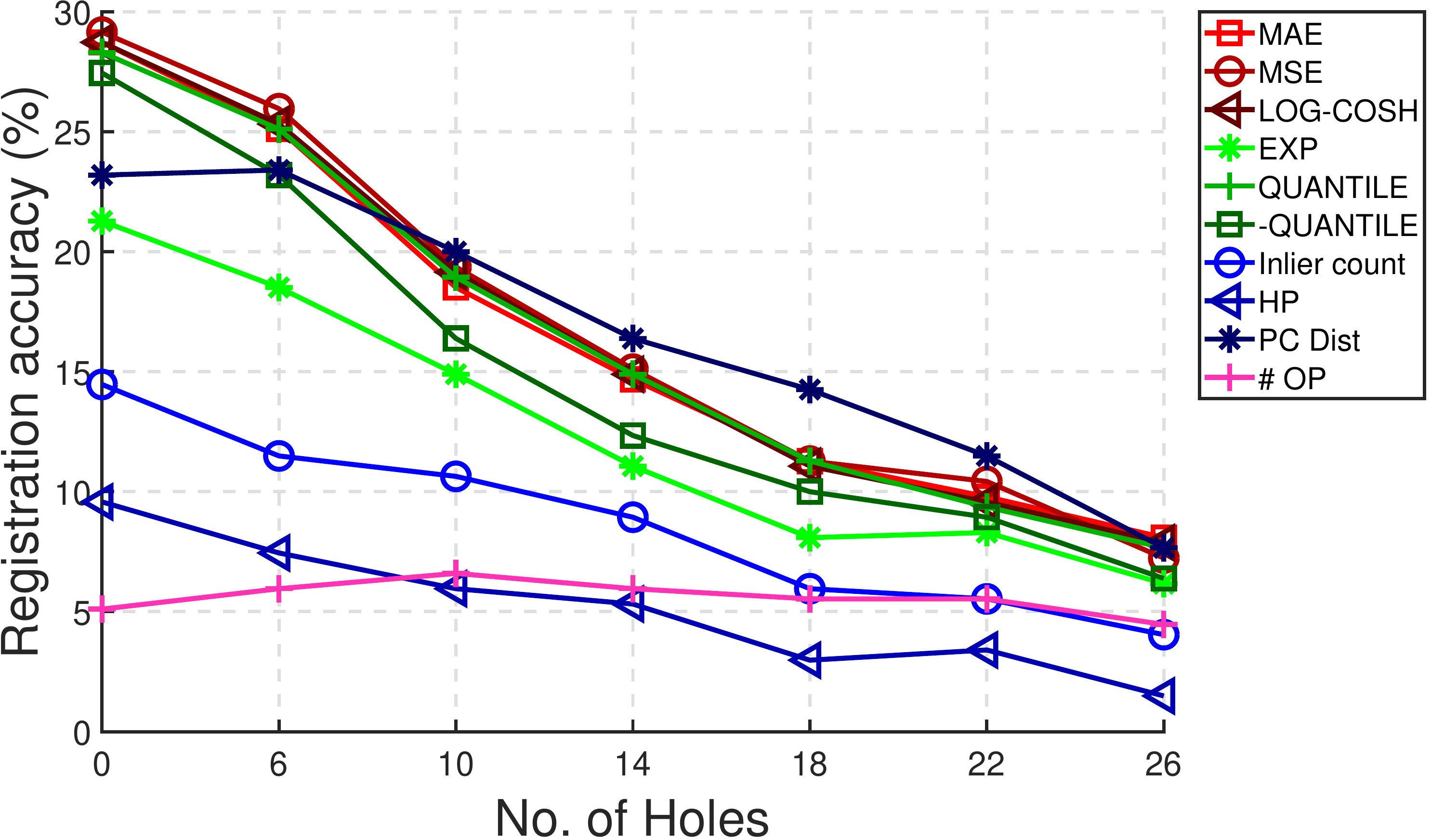}}
	\end{minipage}
	\begin{minipage}{0.49\linewidth}
		\raggedleft
		\subfigure[]{
			\includegraphics[width=0.9\linewidth]{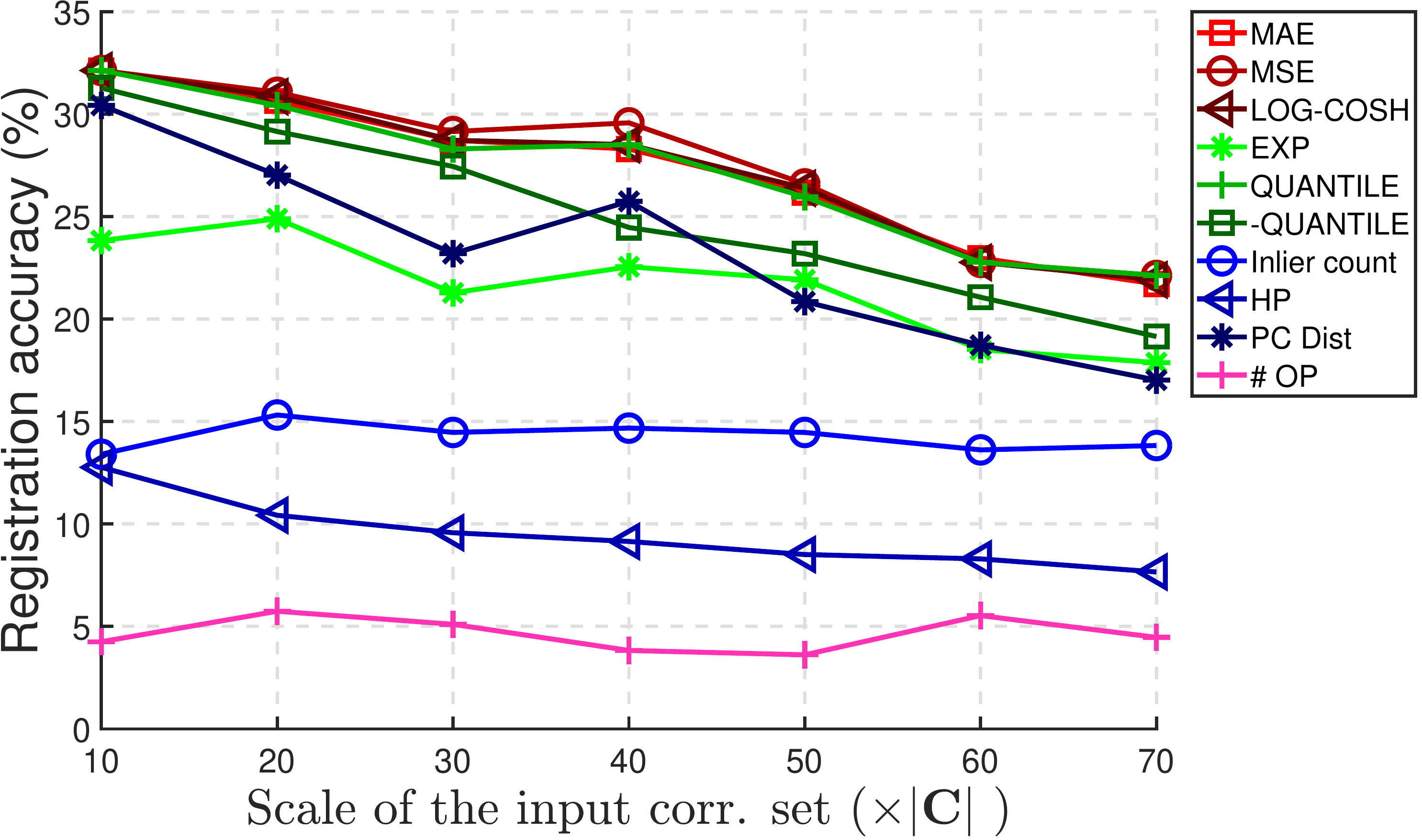}}
	\end{minipage}
	\hfill
	\begin{minipage}{0.49\linewidth}
		\raggedright
		\subfigure[]{
			\includegraphics[width=0.9\linewidth]{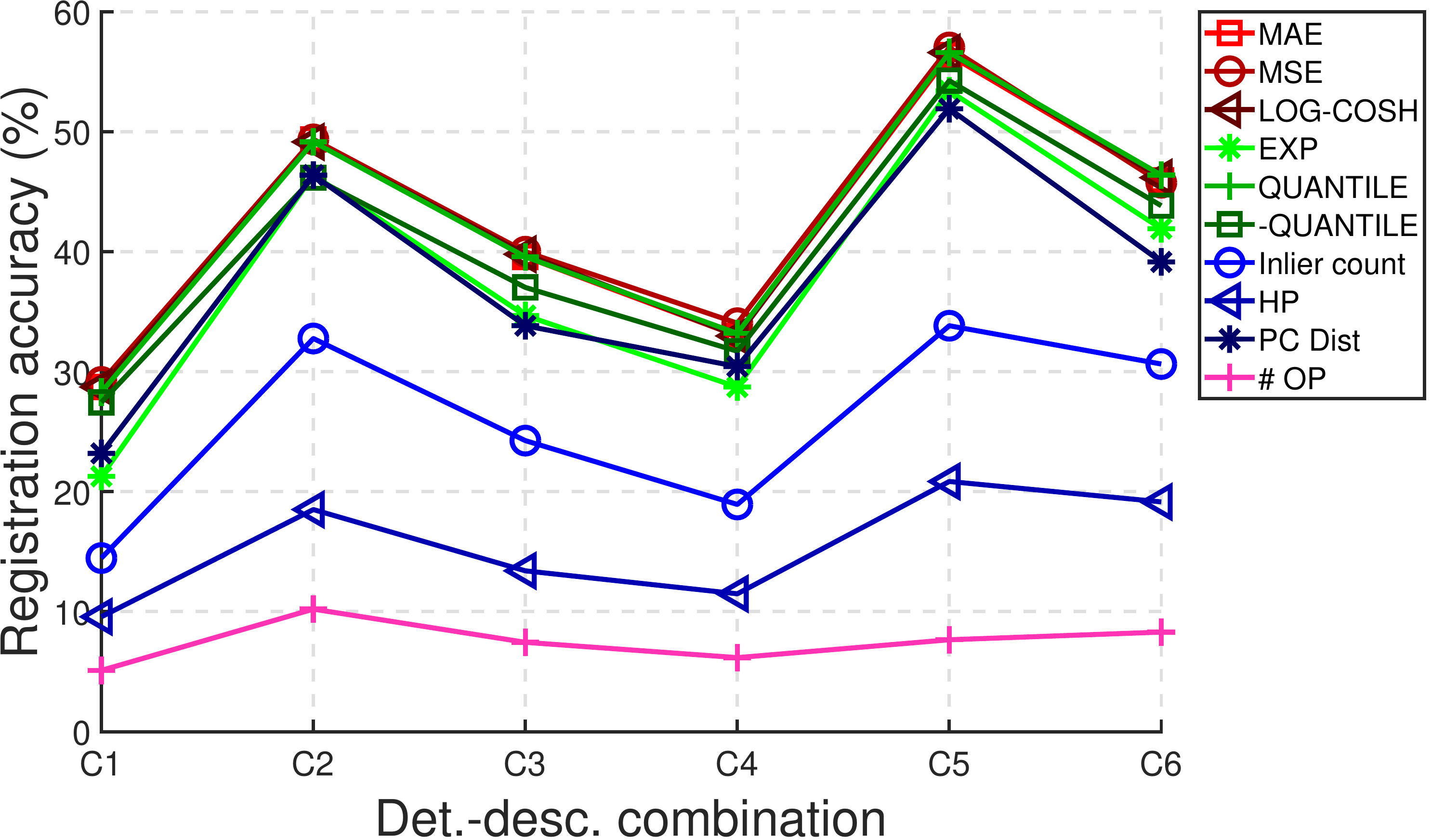}}
	\end{minipage}
	\caption{3D rigid registration accuracy performance of tested metrics with respect to different experimental configurations.}
	\label{fig:robust}
\end{figure*}

{\bf First}, the proposed metrics (except for EXP) and PC Dist generally surpass other tested metrics on all datasets. In particular, our metrics deliver better performance than PC Dist especially on the U3M dataset when $d_{rmse}$ is smaller than 2.5 pr. It suggests that our metrics often produce more accurate registrations than compared metrics. Remarkably, the four datasets have different application scenarios and data modalities. Even though, the proposed metrics consistently achieve top-ranked performance on cross-dataset experiments.

{\bf Second,} among our proposed metrics, MAE, MSE, and LOG-COSH behave better than QUANTILE, -QUANTILE and EXP. We can infer that {\bf 1)} the effect of using different continuous functions to measure the contributions of inliers on the eventual registration performance is not obvious, as long as treating all outliers equally; {\bf 2)} assigning either positive or negative weights to outliers can slightly degrade the registration performance; {\bf 3)} although EXP also assigns larger weights to more accurate inliers, the gap between accurate inliers and inaccurate ones is smaller than those of MAE, MSE, and LOG-COSH, which can result in performance deterioration as well.

{\bf Third,} hypothesis evaluation is critical to RANSAC estimators. Using different hypothesis evaluation metrics can result in dramatic registration performance variation. For instance, the best metric outperforms the worst metric by about 50 percents in terms of registration accuracy on the U3OR dataset when $d_{rmse}$ equals 5 pr. This highlights the importance of hypothesis evaluation to accurate registration and demonstrates the significance of this research.

We present some visual results in Fig.~\ref{fig:visualization}, which displays the point-wise registration errors by MSE, inlier count, and \# OP metrics and the registration results by MSE. The qualitative results are generally consistent with quantitative results.
\subsubsection{Results with Different Configurations}\label{subsubsec:diff_conf}
Many factors will affect the quality of input correspondences to RANSAC estimators. Specifically, we examine the following factors.
\\\\\noindent\textbf{Gaussian noise} Six levels of Gaussian noise with different standard deviations (0.5 pr to 5.0 pr with a gap of 0.5 pr) are injected to the target point cloud.
\\\\\noindent\textbf{Uniform data decimation} We uniformly reject approximately 80\%, 60\%, 40\%, 30\%, 20\%, and 10\% points from the target point cloud to change the sparsity of data to be aligned. 
\\\\\noindent\textbf{Random data decimation} Different from uniform data decimation, we randomly reject 80\%, 60\%, 40\%, 30\%, 20\%, and 10\% points from the target point cloud. This additionally causes data non-uniformity.
\\\\\noindent\textbf{Holes} We add 6, 10, 14, 18, 22, 26 holes to the target point cloud, respectively. We synthesize a hole by first performing $k$-nearest-neighbor search for a point in the point cloud and then removing a portion of neighboring points.
\\\\\noindent\textbf{Varying scales of the input correspondence set} We first order  initial correspondences based on Lowe's ratio score~\cite{lowe2004distinctive}. Then, the number of input correspondences to RANSAC estimators varies from 10\%$\times {\bf C}$ to 70\%$\times {\bf C}$ with a step of 10\%$\times {\bf C}$.
\\\\\noindent\textbf{Varying detector-descriptor combinations}
Two 3D keypoint detectors, including H3D and intrinsic shape signatures~\cite{zhong2009intrinsic}, and three local geometric descriptors including SHOT, spin images (SI)~\cite{johnson1999using}, and local feature statistics histograms (LFSH)~\cite{yang2016fast} are considered to generate initial correspondences. All combinations are examined, i.e., H3D+SHOT (C1), H3D+SI (C2), H3D+LFSH (C3), H3D+SHOT (C4), H3D+SI (C5), and H3D+LFSH (C6).

We experiment with the U3M dataset. The results in terms of the inlier ratio and the number of inliers under above experimental configurations are presented in Fig.~\ref{fig:nuisance_info}. We can see that inputs with a variety of inlier ratios and inlier counts are generated. The results under different experimental configurations are shown in Fig.~\ref{fig:robust}. We can make the following observations.

{\bf First}, the proposed metrics (except for EXP) achieves top-ranked performance under all tested conditions. This demonstrates that the proposed metrics are robust to common nuisances such as Gaussian noise, data decimations, and holes. Moreover, within the pipeline of local-feature-matching-based 3D rigid registration, other modules such as keypoint detection, local feature descriptor, and correspondence selection, can also affect the performance of RANSAC estimators. Even though, our metrics consistently achieve superior performance.

{\bf Second}, the PC Dist metric outperforms ours under the influence of holes. It is because holes may significantly change the geometry information of a point cloud, and in this case leveraging the overall point clouds for hypothesis verification is more reasonable. Nonetheless, we can see that the gap between the curve of PC Dist metric and ours is not obvious. In addition, leveraging the whole point clouds for hypothesis verification is ultra time-consuming as will be verified next.
\subsubsection{Time Efficiency Results}
The average time costs of RANSAC estimators with different hypothesis evaluation metrics for registering one point cloud pair in each dataset are reported in Fig.~\ref{fig:time}.

It is salient that RANSAC estimators with point-cloud-based metrics are dramatically more time-consuming than those with correspondences-based metrics. This is because point-cloud-based metrics need the whole point clouds for hypothesis evaluation while correspondences-based ones only need sparse keypoints. In addition, point-cloud-based metrics need establishing corresponding relationship between points again, which requires nearest neighbor search. Therefore, although PC Dist can also achieve accurate and robust registrations in some test conditions, it is not practical in fact due to its low efficiency. 

\section{Conclusions and Future Work}\label{sec:conc}
This paper investigated the problem of defining efficient and robust metrics for RANSAC hypotheses and 3D rigid registration. We first analyzed the contributions of inliers and outliers, and then proposed several efficient and robust metrics with different designing motivations for RANSAC hypotheses. The technique details are simple while the performance boosting is impressive. Experiments were conducted on datasets with different application scenarios, data modalities, nuisances, and experimental configurations. This generates inputs with various inlier ratios, scales, and spatial distributions, thus ensuring a comprehensive experimental evaluation for proposed metrics. Moreover, all existing metrics for RANSAC hypotheses (to the  best of our knowledge) were compared and we showed that our metrics can achieve more accurate registrations while being ultra efficient. Besides, our metrics can adapt to other RANSAC variants and is quite robust to parameter changes.

 The experimental results have revealed the following findings:

\begin{figure}[t]
	\centering
	\includegraphics[width=1\linewidth]{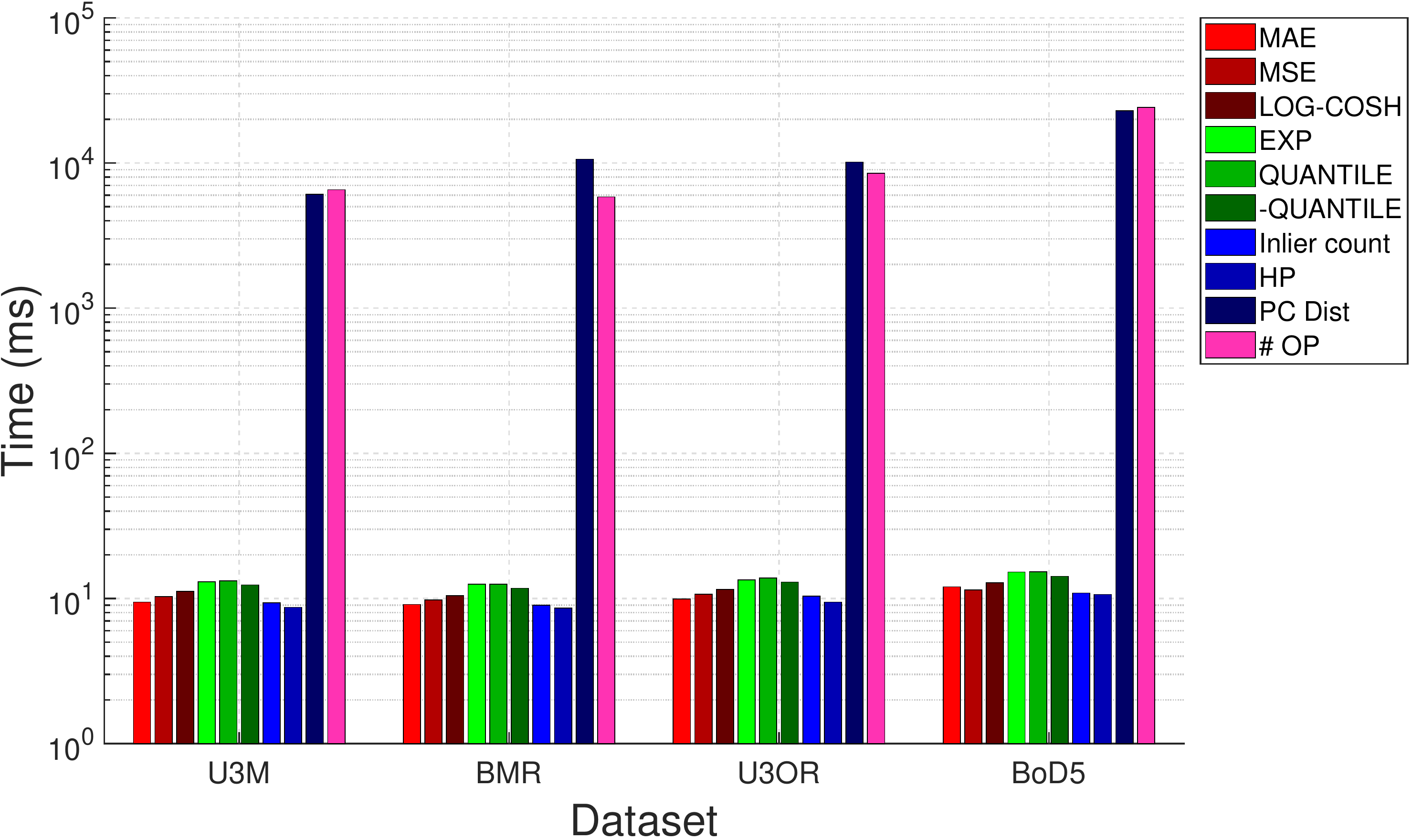}\\
	\caption{Average time costs of RANSAC estimators with different hypothesis evaluation metrics for registering a point cloud pair on four experimental datasets.}
	\label{fig:time}
\end{figure}
\begin{itemize}
	\item  {\bf Significance of hypothesis evaluation metrics.} We stress that the problem of defining proper hypothesis evaluation metrics for RANSAC has been overlooked. Our experimental results demonstrate that using different hypothesis evaluation metrics can result in significant registration variation. For instance, our metrics can improve the registration accuracy by more than 20 percents on the U3M and U3OR datasets. Moreover, the proposed metrics can also adapt to other RANSAC variants and bring performance improvement. Therefore, this paper may draw future research attentions from the community on this problem.
	\item {\bf Not all inliers are equal, while all outliers should be equal.} It is a new conclusion drawn by this work, which is supported by comprehensive experiments. We show that metrics following this rule can achieve a good balance among accuracy, robustness, and efficiency. In addition, these metrics are ultra stable when changing parameters. This is a critical trait, indicating that parameter tuning works when changing applications, datasets, and other parameters of RANSAC can be avoided. Therefore, the proposed metrics are quite useful in practical applications.
	\item {\bf Function design rules for inliers.} Our experimental results find that the behaviors of MAE, MSE, and LOG-COSH metrics are similar, although different continuous math functions are employed. The common feature of them is that they are continuous monotonic decreasing functions in the range of $[0,1]$ with respect to correspondence transformation error $e({\bf c})$.
	\item {\bf Correspondences-based metrics can achieve accurate registrations.} Existing metrics for RANSAC hypotheses in the context of 3D rigid registration are either correspondences-based or point-cloud-based. In prior works, point-cloud-based metrics, which incorporate global context information, are shown to behave better than correspondences-based ones under some particular conditions. However, our work shows that correspondences-based metrics can achieve even better performance when carefully weighing up the contributions of inliers and outliers. Moreover, correspondences-based metrics are at least three orders of magnitude faster than point-cloud-based ones.
\end{itemize}

Since the proposed metrics are demonstrated to be fast, accurate, and robust to nuisances and parameter changes, we expect applying them to practical applications relying on 3D rigid registration such as LiDAR SLAM, 3D change detection, and 3D object recognition.

\section*{Acknowledgments}
The authors would like to thank the publishers of datasets used in our experiments for making them publicly available. 
\bibliographystyle{IEEEtran}
\bibliography{mybibfile}

\end{document}